    \documentclass[pdflatex,sn-mathphys-num]{sn-jnl}
    
    \usepackage{graphicx}%
    \usepackage{multirow}%
    \usepackage{amsmath,amssymb,amsfonts}%
    \usepackage{amsthm}%
    \usepackage[title]{appendix}%
    \usepackage{xcolor}%
    \usepackage{textcomp}%
    \usepackage{manyfoot}%
    \usepackage{booktabs}%
    \usepackage{algorithm}%
    \usepackage{algorithmicx}%
    \usepackage{algpseudocode}%
    \usepackage{listings}%
    \usepackage{bookmark}
    
    \usepackage{lmodern}
    \usepackage{comment}
    \usepackage{makecell}
    \usepackage{pifont}
    
    \newcommand{\xmark}{\ding{55}}%
    
\begin{document}
    
    \title[Article Title]{Discovering Multiscale Deep Formulas in Complex Systems via Neural-Guided Lambda Calculus}
    
    \author[1,2]{\fnm{Hanqiao} \sur{Yu}}
    \equalcont{These authors contributed equally to this work.}
    
    \author*[1,2]{\fnm{Shusen} \sur{Yang}}\email{shusenyang@mail.xjtu.edu.cn}
    \equalcont{These authors contributed equally to this work.}
    
    \author*[1,3]{\fnm{Xuebin} \sur{Ren}}\email{xuebinren@mail.xjtu.edu.cn}
    
    \author[1,2]{\fnm{Cong} \sur{Zhao}}

    \affil*[1]{\orgdiv{National Engineering Laboratory for Big Data Analytics},
    \orgname{Xi'an Jiaotong University}, 
    \orgaddress{\city{Xi'an}, \postcode{710049}, \state{Shaanxi}, \country{China}}}
    \affil*[2]{\orgdiv{School of Mathematics and Statistics},
    \orgname{Xi'an Jiaotong University}, 
    \orgaddress{\city{Xi'an}, \postcode{710049}, \state{Shaanxi}, \country{China}}}
    \affil*[3]{\orgdiv{School of Computer Science and Technology, Faculty of Electronic and Information Engineering},
    \orgname{Xi'an Jiaotong University}, 
    \orgaddress{\city{Xi'an}, \postcode{710049}, \state{Shaanxi}, \country{China}}}
    
    
    \abstract{A fundamental problem in science is identifying underlying patterns of complex systems in the form of concise mathematical formulas. Current Artificial Intelligence (AI)-based methods have shown strong performance in single-scale systems, yet remain limited in identifying scale-specific formulas in multiscale complex systems. We present Deflex, an end-to-end AI method to automatically extract multiscale formulas with potentially different forms, including invariants and distributions, from complex systems. Deflex consists of two subsystems named Deflexformer and Deflexpressor. Deflexpressor is a lambda-calculus symbolic regression model for higher-order formulas. Deflexformer is a decomposable deep energy model for learning unified representations across scales. Deflexpressor generates synthetic data to pre-train Deflexformer, which then guides formula discovery by decoupling multiscale latent relationships. Across six representative complex systems with diverse behaviors, Deflex achieves up to 7-fold higher efficiency than the state-of-the-art methods while enabling automated multiscale discovery. Our work could be a useful tool for scientific discovery across disciplines.}
    
    
    
    
    \maketitle
    
    \section{Introduction}\label{sec1}
    Discovering rules in mathematical forms, such as energy conservation and Boltzmann distribution, in nature and society has always been an important endeavor in scientific research~\cite{wigner1990unreasonable,doi:10.1126/science.1165893}.
    In recent years, artificial intelligence (AI) has played an increasingly significant role 
    in identifying mathematical formulas from different systems~\cite{doi:10.1126/science.1165893,irrgang2021towards,doi:10.1073/pnas.1917285117,varadi2022alphafold}. 
    From computational and cognitive perspectives, AI-based automatic formula discovery for complex systems has to solve three fundamental scientific problems:
    1) the volume challenge, indicating that the huge number of interacting elements would result in computational and representational prohibitiveness of formula findings,
    2) search-space explosion, meaning that all possible free-form formulas increase exponentially as the number of variables and potential expression depth increase, 
    due to the volume and complex nonlinear interactions in complex systems~\cite{udrescu2020ai,doi:10.1126/sciadv.aay2631,irrgang2021towards,doi:10.1073/pnas.1917285117, bellomo2011modeling}, and 
    3) the scale gap issue, suggesting that the mathematical rules emerge at different scales in complex systems and automatic discovery requires unified representations and analysis at different levels~\cite{bar2002general, national2012national}. 
    Facing the above challenges, discovering genuine mathematical formulas in a fully automated fashion, 
    without pre-specified equation structures, 
    remains largely limited to single-scale and few-body systems of lower complexity~\cite{fortunato2018science,pomeau2016long,wigner1990unreasonable,udrescu2020ai,10.5555/3495724.3497186,doi:10.1126/sciadv.1602614, gao2024learning}.

    Meanwhile, increasing effort has been put into research on complex systems, such as fluid dynamics~\cite{doi:10.1073/pnas.1917285117}, crowd movement~\cite{van2018continental}, climate evolution~\cite{national2012national,hasselmann1976stochastic}, and social systems~\cite{bellomo2011modeling}, which explain many complex phenomena and behaviors in the real world. 
    Many distinct properties of complex systems come from the transition to chaos and the emergence of order from disorder across different spatial and temporal scales~\cite{vicsek2012collective,kaneko2001complex}. 
    Such chaos and order emergence also introduces different levels of regularity, requiring scientific analysis to account for distinct mathematical forms: deterministic descriptions in conservation laws and probability distributions for stochastic behaviors~\cite{weinan2011principles}.
    AI methods have demonstrated great potential for discovering the governing rules behind the interactions of massive elements at multiple scales in these complex systems. 
    For instance, their capacity to efficiently approximate solutions for complex known equations in physics~\cite{pun2019physically}, 
    their ability to discern intricate patterns from large-scale observational data in domains such as meteorology or climate science~\cite{bi2023accurate}, 
    and the emerging success of large models in learning or representing complex mathematical relationships from scientific data~\cite{romera2024mathematical} all underscore this potential.
    
    Unfortunately, no existing AI methods can achieve automatic discovery of multiscale deep formulas in complex systems. 
    While methods like sparse regression can effectively learn mathematical patterns from large datasets~\cite{doi:10.1073/pnas.1517384113,10.5555/3495724.3497186}, their reliance on pre-defined functional forms limits the exploration of unknown regularities. 
    Symbolic regression (SR), based on symbolic trees and genetic programming~\cite{koza1994genetic, doi:10.1126/science.1165893, 10.5555/3495724.3497186, doi:10.1126/sciadv.aay2631, udrescu2020ai}, is a powerful tool for automatically discovering mathematical formulas with unknown forms from data.
    However, traditional SR methods suffer from low success rates and computational efficiency for identifying complex relationships.
    In particular, discovering probability distributions from raw observational data—rather than merely fitting predefined distribution functions—represents a core challenge for multiscale formula discovery, as stochastic patterns naturally emerge at coarse-grained scales~\cite{weinan2011principles}.
    While some advanced methods (e.g., AI Feynman~\cite{udrescu2020ai}) can handle distribution fitting, they require substantial manual specification of candidate functional forms.
    Recently, substantial progress has been made in symbolic regression, which has greatly enhanced the discovery of laws in real-world data. For instance, physics-inspired symbolic regression has explored incorporating physical knowledge (e.g., higher-order symmetries) as priors~\cite{keren2023computational, sun2023symbolic}. Also, neural network-integrated SR discovers novel patterns from representations learned from deep models (e.g., GNNs)~\cite{10.5555/3495724.3497186}. 
    Besides, many studies have enabled symbolic regression to find complex relationships on benchmark datasets by improving the searching strategies with evolutionary algorithms~\cite{10.1609/aaai.v38i11.29187,udrescu2020ai}, 
    reinforcement learning~\cite{NEURIPS2021_d073bb8d}, or Monte Carlo tree search~\cite{sun2023symbolic}.
    Nonetheless, without higher-order representations and tailored design, the automatic discovery of the above studies can hardly work for multiscale complex systems.

    In this article, 
    we propose a novel framework of \underline{De}ep \underline{F}ormula discovery for Comp\underline{lex} systems (Deflex) by incorporating new symbolic regression systems with strong expressiveness, 
    unified formulation methodology, 
    and decomposable neural network architecture.
    The main contributions of our work are summarized as follows:
    
    We first regard all mathematical formulas as probability distributions~\cite{siegenfeld2020introduction}, which can unify the system patterns at different scales.
    Specifically, equations and invariants are treated as distributions with very narrow peaks,
    which indicates their tiny errors or uncertainty. 
    Original distribution-based formulas are treated as distributions with wider peaks. 
    We then formulate these distribution patterns through energy-based models (EBMs), where each pattern is represented as an energy function. 
    This formulation facilitates both mathematical modeling and algorithmic processing of these patterns. Particularly, by measuring the likelihood of EBMs over the observation data, we build a tractable learning framework for finding the optimal energy functions. 
    
\begin{figure*}[p!]
    \centering
    \includegraphics[width=\linewidth]{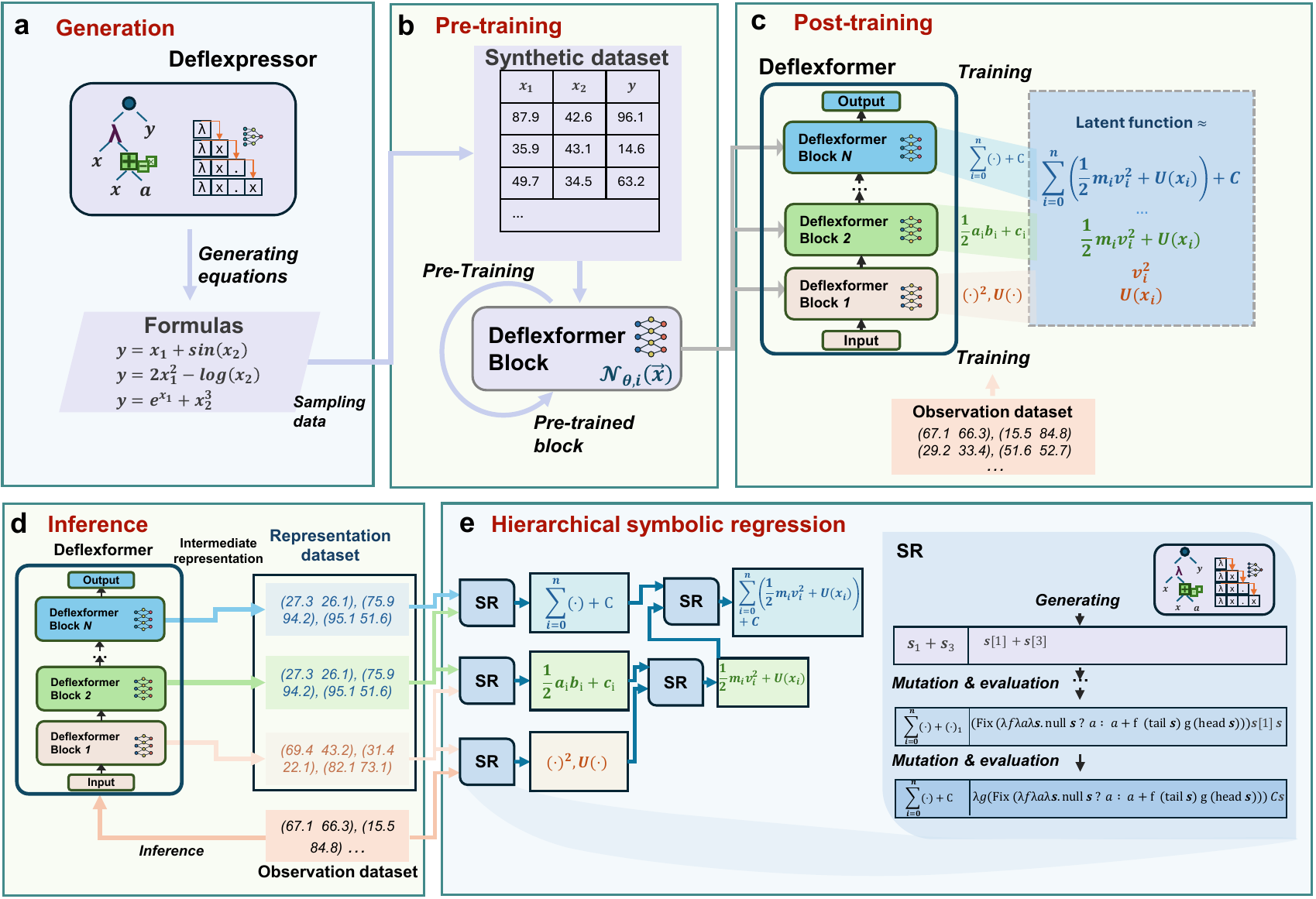}
    \caption{
        Illustration of the workflow in Deflex.
        Deflex utilizes two subsystems: Deflexpressor and Deflexformer,
        where the former is a symbolic regression subsystem based on lambda calculus and the latter is an energy-based DNN subsystem.
        The overall process flows sequentially from \textbf{a} through \textbf{b}, \textbf{c}, \textbf{d}, and finally to \textbf{e}.
        \textbf{a}, Deflexpressor generates a large number of expressions,
        from which it samples a synthetic dataset.
        \textbf{b}, the synthetic dataset is used to pre-train a Deflexformer block,
        a Transformer-like neural network.
        \textbf{c}, the pre-trained block is repeatedly cascaded multiple times to build Deflexformer,
        which is then trained to learn latent formulas on the observation dataset in the target complex system.
        \textbf{d}, the trained Deflexformer is fed with the observation dataset and some random data for inference,
        which generates a large number of data representations output by all blocks.
        \textbf{e}, the multi-level representation dataset from all Deflexformer blocks is processed by Deflexpressor through hierarchical symbolic regression.
        The SR framework iteratively generates expression candidates,
        evaluates them across different representation levels,
        and refines them through mutation to discover the desired formulas as both mathematical expressions and executable lambda-calculus code.
    }
    \label{fig:deflex}
    \end{figure*}

    With these unified representations and formulations, the overall workflow of Deflex is shown in Fig. \ref{fig:deflex}.
    We next develop Deflexformer, the neural network subsystem of Deflex.
    Deflexformer is a deep neural network (DNN) for learning the energy functions. It is designed as a self-attention-based network (Fig. \ref{fig:arch}a,b) to leverage the demonstrated efficacy of self-attention architectures in processing set-structured data and capturing the patterns for both element-wise and global variables. Different from existing neural-integrated methods, Deflexformer can further decompose the learned functions into elementary components that are tractable for symbolic regression. Deflexformer is composed of multiple Transformer-like blocks, each of which represents a distinct elementary mathematical transformation. Thus, any intricate mathematical relationships learned by Deflexformer, can be collectively parameterized by the modular transformations of all Deflexformer blocks.
    The neural network with decomposable blocks can efficiently guide symbolic regression algorithms for tractable learning of mathematical formulas from each component, rather than from the whole.
    
    However, traditional tree-based symbolic regression may have significant limitations in both expression ability and discovery efficiency for higher-order mathematical relationships (such as summation or mapping) over a large volume of interacting elements in complex systems.
    The former limitation is lacking native support for variable bindings, so a standard expression tree must either rely on a rigid, predefined summation primitive or force the structure to unroll into an intractably long arithmetic chain of pairwise additions. The latter is due to the combinatorial search space of potentially massive arguments for complex systems.
    To address these limitations, we present Deflexpressor, a symbolic regression subsystem that leverages lambda-calculus expressions.
    The inherent capabilities of lambda calculus for function abstraction, variable binding, and higher-order functions provide a powerful and flexible representational framework.
    This allows Deflexpressor to express and discover free-form higher-order mathematical relationships, such as mapping and aggregating interactions over distinct pairs without pre-specifying the operators.
    In addition, Deflexpressor can efficiently navigate the search space for complex symbolic structures via two novel designs. One is to guide lambda expression construction by incorporating a rule-based generator and an auto-regressive neural network. The other optimizes the efficiency-diversity trade-offs of expression generation and mutation by extending genetic programming mutations with transformations specified for lambda calculus (e.g., inspired by beta-reduction or exploring variations in function arity and application).
    
    The workflow of our approach proceeds as follows.
    First,
    Deflexpressor generates a large number of formulas for sampling a synthetic dataset (Fig. \ref{fig:deflex}a).
    The synthetic dataset is then used to pre-train a Deflexformer block,
    which is a Transformer-like neural network (Fig. \ref{fig:deflex}b).
    This pre-training serves as prior knowledge for the subsequent step,
    where we cascade the multiple copies of the pre-trained block as the whole Deflexformer to learn patterns from the observation dataset of a certain complex system (Fig. \ref{fig:deflex}c).
    During learning,
    Deflexformer employs element-wise and time-wise self-attention mechanisms to capture the interactions among different entities and time frames.
    Each trained block then essentially represents a distinct component of the latent formulas of the patterns.
    In this phase,
    the observation dataset,
    together with some random data,
    is then fed to the trained Deflexformer for inference,
    which generates a large volume of data representations output by all blocks (Fig. \ref{fig:deflex}d).
    Finally, the representation dataset output by the elementary blocks in Deflexformer is fitted using Deflexpressor to obtain the desired formulas from each component,
    through iterative formula generation and mutation in the form of lambda expressions (Fig. \ref{fig:deflex}e).
    
    Thus far, we have outlined the design principles and general processes of the proposed method, which aim to address the aforementioned challenges for AI-based formula discovery from complex systems.
    Concerning the volume problem, Deflex uses a symbolic regression system based on lambda calculus to compress the mathematical rules into concise forms,
    and utilizes scalable deep learning and symbolic regression algorithms to handle large-scale datasets.
    For the complexity of interactions, we design a self-attention-based DNN for Deflex to decompose the complex systems into different scales and sub-problems, making the approach efficient for highly complex formulas.
    For the scale gap problem, Deflex uses energy-based models to unify invariant and distributional patterns across scales.
    A detailed illustration of the system modelling, the subsystems design, and the specific workflow will be presented in the Method Section.
    
    \section{Results}\label{sec3}
    We evaluate the effectiveness and efficiency of Deflex in six representative complex systems with different behaviors. 
    Particularly, these systems can be categorized into simulation experiments, including particle motion, two-phase water particles, and fluid dynamics, and the real-world data of human and bird collective movements. More details about the experiments can be found in the SI.
    
    \subsection{Compared methods}
    We benchmark our approach against several leading AI methods for formula discovery, 
    including state-of-the-art SR method Operon~\cite{10.1145/3377929.3398099}, 
    typical physics-inspired SR method SciMED~\cite{sun2023symbolic}, leading sparse regression method SINDy~\cite{doi:10.1073/pnas.1517384113}, 
    and AI Feynman 2.0~\cite{udrescu2020ai}, 
    as well as classic SR methods gplearn~\cite{Stephens2025trevorstephens}, PySR~\cite{cranmerInterpretableMachineLearning2023}
    and DEAP~\cite{DEAP_JMLR2012} in terms of accuracy, 
    efficiency, and adaptability to multiscale systems.
    
    Different from Deflex with energy models and lambda calculus, existing AI methods (mainly based on symbolic regression) often support limited forms of problems and formulas.
    To maximize comparability, we made several accommodations for these compared methods.  
    For the comparatives (PySR, AI Feynman, and DEAP) lacking native support for higher-order operators (e.g., map or sum), we added these operators to their symbol set manually.
    For the rest ones (Operon, SciMED, SINDy, and gplearn) that are incompatible with these operators, we provided an alternative. That is, if such an operator was at the top level of the target formula, we applied the method for the sub-expressions only and then handled the top-level operators within the loss function manually.  
    Similarly, for methods unable to approximate distributions, we provided custom energy-based loss functions (see SI for comparison details).
    Note that some of these frameworks have variants that would gain better performance with scientists in the loop (e.g., manually providing feature sets or constraints on the expression tree for SciMED and AI Feynman), while we mainly used the raw version for simplicity despite the above adaptations.
    A method was considered to have failed a task if it still could not discover the rule after these best-effort accommodations.
    
    \subsection{Evaluation metrics}
    Since uncovering formulas within a complex system may represent an open-ended challenge, 
    we incorporate qualitative assessments, basic quantitative analyses, 
    renormalization and coarse-graining techniques~\cite{villegas2023laplacian} in the performance evaluation.
    
    For established or empirical formulas, we first qualitatively assess whether the structure and parameters of the discovered formulas align with the ground truth. The formulas of invariants are considered successful recovery if they can be expressed as the correct formula with an additional stochastic residual term. Table \ref{tab1} lists all discovered formulas by Deflex in the studied six systems, compared with the ground truths. Note that, for undocumented discoveries, we provide justifications for their possible real-world interpretations. For reference, we also summarize the capability of the compared methods for discovering these formulas in Table 2. 
    
    To quantitatively evaluate the accuracy of discovered formulas, we consider two cases based on the different forms of formulas separately.
    \begin{itemize}
        \item For the formulas in the form of equations and invariants, we employ the relative mean square error (RMSE) between the prediction and the true value. Specifically, all compared methods are given the same convergence threshold. Note that, in such a case, the RMSEs of different methods are hardly comparable since there is almost no structural difference (except for residual terms with negligible errors) in the commonly identified formulas in these forms.
        \item For distribution-based formulas, we use the negative log-likelihood (NLL) loss defined in Eq. (\ref{eq:nll}) in the method section, which measures the likelihood of the distribution fitting real data. The feasible methods were also set with the same convergence constraint. Note that we also report the RMSE to measure the fitting error of simple distributions for reference.
    \end{itemize} 
    Fig. \ref{fig:analysis_ab}a compares the NLL loss of Deflex's distribution discovery in comparison with other methods, and reports the RMSE on the commonly identified equations and invariants. Besides, we compare the recovery rate to empirically measure the percentage of times certain formulas can be consistently recovered from the data.
    
    To evaluate the efficiency of Deflex, we run Deflex and other comparative algorithms on a workstation with 32 CPU cores working in parallel, 
    and two Tesla V100 GPUs for GPU-accelerated computing (details shown in SI Sec. 3). 
    We compare the time efficiency of Deflex with that of other methods by recording the run-time of finding the same or equivalent formulas from the same dataset, within the same tolerable error in NLL or RMSE.
    Note that the run-time of Deflex is measured from applying Deflex on the observation dataset, including post-training and inference of Deflexformer, and hierarchical symbolic regression for formula discovery. It excludes the time cost of the Deflexformer block pre-training, which was conducted offline over 10,000 sampled pairs and independent of any studied system.
    A discovery is considered successful when its error remains within a small margin of the residual error,
    defined as the minimum achievable discrepancy when fitting the ground truth formula to observations~\cite{udrescu2020ai}.
    This baseline accounts for irreducible noise from measurements and stochastic fluctuations in the system (details in SI).
    Fig. \ref{fig:analysis_ab}b shows the comparison results.
    
    In our experiments,
    we employ different observation scales by varying the spatiotemporal scope of data sampling.
    Small-scale observations use localized sampling (e.g., selecting spatiotemporally neighboring data points) to capture fine-grained interactions and facilitate the discovery of governing equations,
    while large-scale observations employ uniform random sampling across the entire domain to reveal macroscopic statistical patterns such as distributions (see SI for detailed sampling protocols).
    
    One of the most important characteristics of complex systems is the relevance between scales and behaviors.
    For evaluating the effectiveness of multiscale formula discovery, we introduce the techniques of coarse-graining and renormalization~\cite{villegas2023laplacian} in our experiments. 
    Coarse-graining refers to grouping the small (fine-grained) system elements $\{e_f\}$ into a set of large (coarse-grained) entities $\{e_c\}$, 
    e.g., aggregating adjacent particles into clusters or merging smaller fluid grid cells into larger ones. 
    Renormalization means transforming the formula $P(\{e_f\}, J)$ with parameters $J$ describing a system at fine-grained level elements $\{e_f\}$, 
    into the formula $P(\{e_c\},J')$ in the same form, with the only difference in parameters $J'$ describing coarse-grained level entities $\{e_c\}$. 
    Scale-invariant and self-similar systems often satisfy formula renormalization across multiple different scales. 
    In our experiments, we also set up the experimental environments for water molecules 
    and fluid dynamics to evaluate the performance of our Deflex framework on such systems. 
    Specifically, we measure the distance between our discovered formulas at different scales with the normalized ground-truth formulas at corresponding scales.
    Here, since all the formulas discovered by Deflex can be represented in the probabilistic form, we use Earth Mover's Distance (EMD)~\cite{villani2009optimal} as an alternative error measure. EMD is particularly useful for distributions with very sharp peaks and small overlaps.
    \begin{equation}
        EMD(\mu, \nu)=\inf _{\gamma \in \Gamma(\mu, \nu)}\left(\mathbf{E}_{(x, y) \sim \gamma} d(x, y)^p\right),
    \end{equation}
    where $\mu$ and $\nu$ are two distributions, 
    $\gamma$ is the set of all possible joint distributions of $\mu$ and $\nu$, and $d(x, y)$ is the distance between $x$ and $y$.
    Fig. \ref{fig:analysis_cd}a compares the EMD value across scales in the scenarios of water particle motion and fluid dynamics.
    
    \subsection{Particle motion}\label{subsec1}
    
    The study of the motion of particles, like pollen, molecules, and fundamental particles, 
    has contributed to some breakthrough discoveries,
    such as the law of energy conservation, the Maxwell distribution, the Boltzmann distribution, and the Langevin equation.
    Many of these have helped reveal fundamental understandings of how patterns emerge in systems across scales~\cite{langevin1908theorie, doi:10.1126/science.1165893}.
    For instance, the Langevin equation manifests that small-scale dynamics can be approximated as noise at larger scales.
    
    We first applied Deflex to the particle motion systems of
    rare gases and pollen in water~\cite{langevin1908theorie}. 
    In the rare gas particle motion experiment, we simulated the motion of argon particles at three temperature points near the critical point ($T_c$, $T_c \pm 1$K, where $T_c = 150.72$K) within a cube, which allowed us to capture the system's behavior across different thermodynamic states.
    When the particles approached the cube edge, they experienced a repulsive force analogous to intermolecular forces. 
    See SI 2.1 for a detailed experimental setup.
    In the pollen experiment in water, we simulated water and pollen within a microtube. 
    This represents a typical two-scale system, involving microscopic water molecules and mesoscopic pollen.
    We sampled 16 to 2048 particles for 50 time frames as a data item and trained our network with a dataset of  65,536 sampled data items. 
    For each particle at a time frame, the mass, position, and velocity are embedded into the embedding layer of 64 dimensions.
    
    As shown in Table \ref{tab1}, Deflex successfully identifies a series of formulas, including the energy conservation, Boltzmann distribution, Maxwell-Boltzmann distribution (MBD), MBD with Lennard-Jones potential, and the Langevin equation, with only one training pass of Deflexformer in each of these particle motion systems.
    Notably, since temperature is provided as an input variable in the argon system, the discovered formulas naturally capture the temperature-dependent behaviors across different thermodynamic states, with temperature $T$ appearing explicitly in the energy expressions. 
    In particular, MBD with Lennard-Jones potential and the Langevin equation cannot be discovered by other compared methods in our experiments (shown in Table 2). And those tree-based symbolic regression method requires manual specification
    MBD with Lennard-Jones potential identified by Deflex in the dynamics of rare gas particles is a nonlinear equation that consists of up to 23 nodes, demonstrating Deflex's ability to discover complex nonlinear equations.
    
    On the commonly identified Maxwell-Boltzmann distribution, Deflex manages to achieve the lowest NLL loss among the compared methods in both Argon and Pollen particles (shown in Fig. \ref{fig:analysis_ab}a). 
    Meanwhile, the RMSEs of all compared methods are nearly the same on both energy conservation and momentum conservation in the two systems. This is because the structures of correctly identified formulas are almost identical in these methods, as mentioned before.  
    As shown in Fig. \ref{fig:analysis_ab}b, Deflex incurs significantly less computational time than most of the comparatives. SINDy and Operon are quite close to or sometimes faster than Deflex, but only limited to some lower-order formulas like momentum conservation. This is due to their limitations in formula structures.
    Furthermore, ablation results indicate that our Langevin sampling strategy is essential for efficiently identifying these complex equilibrium distributions (Fig.~\ref{fig:ablation}g).
    
    \subsection{Human and bird mobility}\label{subsec3}
    The movement of human crowds and bird flocks exemplifies significant randomness and multiscale characteristics in complex systems. 
    In the experiment, we used data from the Zurich Carnival dataset~\cite{10.4108/icst.urb-iot.2014.257190} and the eBird bird migration dataset~\cite{eBirdStatusTrends2022} to simulate the collective movement of humans and birds, respectively. 
    These datasets consist of the moving trajectories of humans and birds over a specific period. 
    Specifically, the former dataset contains 24M locations and time points of 29,000 visitors, with 2 minutes as a time frame. 
    We sampled the trajectory data of 16 to 2048 people in 60 time frames (2 hours) for a data snippet in the training stage. 
    We used the reanalysis of 7 species from eBird as the dataset, containing 176M data points. For each data snippet, we sampled 16 individual birds over 256 to 1024 time frames.
    
    As presented in Table \ref{tab1}, Deflex identified multiple formulas, 
    including power law equations~\cite{gonzalez2008understanding}, Lévy flight patterns~\cite{schlapfer2021universal}, and vector navigation laws~\cite{bongiorno2021vector} in human crowds, and the Lévy flight patterns in bird migrations. 
    Besides Deflex, AI Feynman, PySR, and DEAP can also discover the power law distribution and Lévy flight pattern (shown in Table 2). 
    But no other method can identify the vector navigation law.
    It is worth noting that Deflex also uncovered an undocumented power law distribution,
    \begin{equation}
        P\left(\Delta x_t^{(i)}\right) \propto r^{\alpha_0 - \gamma \sum_{j \neq i} \mathbb{I}\left(\left\|x_t^{(j)}-x_t^{(i)}\right\|<2.46\right)}.
    \end{equation}
    This rule may indicate a trend superimposed on the baseline power-law distribution and describe the collective behavior of human crowds: 
    individuals exhibit a higher probability of making a long-distance move as their neighbors do.
    
    For the commonly found rules of the power law distribution and Lévy flight pattern, we first compare the NLL loss and RMSE of Deflex with other methods in Fig. \ref{fig:analysis_ab}a. It shows that Deflex's discovered distributions have consistently high accuracy (both smaller NLL loss and RMSE value) than other capable methods.
    We also compare their time efficiency for discovering these distributions under the same accuracy (i.e., by setting an identical precision threshold for convergence). Fig. \ref{fig:analysis_ab}b shows that Deflex also leads to the least running time among the compared methods.
    
    \subsection{Water particles}\label{subsecwater}
    To further validate the performance of Deflex in multiscale systems, we conducted experiments on cross-scale water particles at different temperatures (300K and 273.15K) and scales. 
    In particular, ice particles can be regarded as either clusters of water molecules or larger particles. 
    This experiment was conducted using the same microtube as the pollen experiment.
    
    For scale-independent characteristics, Deflex demonstrates similar performance in water particles as on Argon and pollen particles, as shown in Table~\ref{tab1}, Figs. \ref{fig:analysis_ab}a and \ref{fig:analysis_ab}b.
    To validate the cross-scale properties, we conducted coarse-graining of water particles by aggregating groups of water molecules that form lattice structures into macro-particles (i.e., ice crystals).
    This process was iteratively repeated to construct a hierarchically coarse-grained model of water. 
    Theoretically, as the coarse-graining progresses, 
    the dynamics of ice crystals were expected to transition from a particle-based description to a Langevin dynamics framework.
    In our analysis, we focused on tracking the short-term dynamics of the water system, typically over a span of 50 time frames.
    This constraint arises because the coarse-grained model cannot capture the formation or melting of ice crystals, 
    limiting its applicability to scenarios where these processes are not significant.
    
    As shown in Fig. \ref{fig:analysis_cd}a, Deflex manages to capture the cross-scale emergence of Langevin dynamics in water particles. 
    In contrast, other methods are limited to observing molecular dynamics at smaller scales and the dynamics of ice crystals at larger scales, failing to bridge the transition across scales.
    It also shows that Deflex manages to detect the scale invariance of the system at the phase transition point of water (i.e., at 273.15K).
    
    \subsection{Fluid dynamics}\label{subsecfluid}
    
\begin{figure}[p!]
        \centering
        \includegraphics[width=\linewidth]{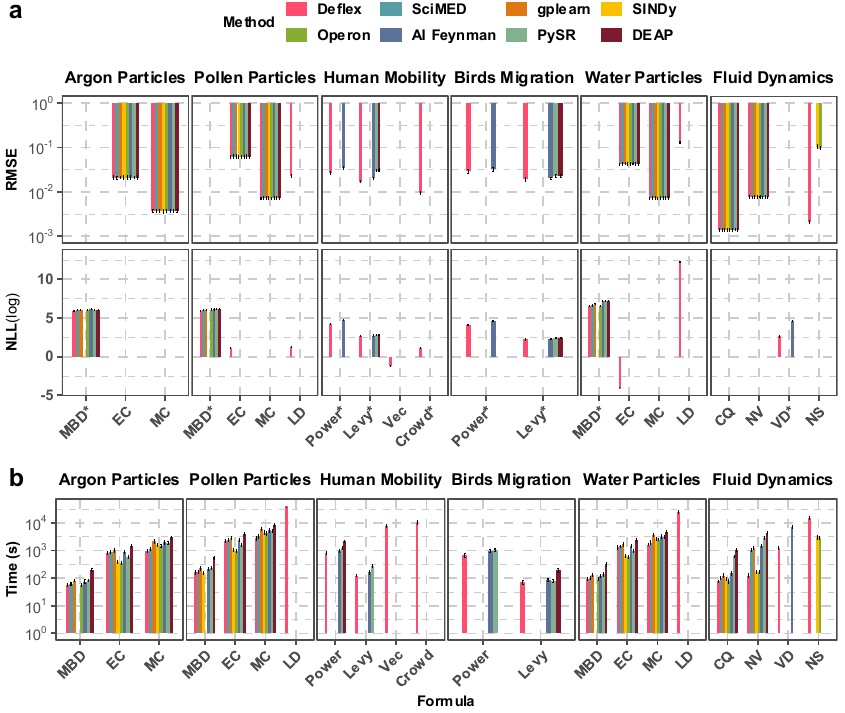}
        \caption{
            Experimental results of Deflex in terms of accuracy (\textbf{a}) and efficiency (\textbf{b}).
            \textbf{a} compares the RMSE and NLL loss of the discovered formulas by Deflex and other methods in the experiments of particle motions,
            human and bird collective movement, water particles, and fluid dynamics.
            The abbreviations represent the Maxwell-Boltzmann distribution (MBD), energy conservation law (EC), 
            momentum conservation (MC), Langevin dynamics (LD), continuity of quantity (CQ), 
            Newton viscosity law (NV), velocity distribution in fluid dynamics(VD), Navier-Stokes equation (NS), 
            power law in human mobility (Power), Lévy flight rule (Lévy), vector navigation law (Vec), and crowd power law, respectively. Among them, the formulas in the form of distributions are marked with $*$.
            The absence of the method indicates that the corresponding metrics are not available for the method.
            \textbf{b} depicts the running time and usability of Deflex compared with other methods in different experiments.
            The horizontal axis represents the running time of the methods and the absence of the method indicates it is infeasible to discover the formula.
            Only Deflex can discover all the formulas in the experiments,
            and the time consumption of Deflex is significantly lower than that of most other methods for complex formulas.
            In panels \textbf{a} and \textbf{b}, error bars show mean $\pm$ standard deviation.
    }\label{fig:analysis_ab}
\end{figure}

    We conducted experiments on fluids (Fig. \ref{fig:analysis_ab}a, b), 
    one of the most typical complex systems in physics~\cite{solera2024beta}. 
    Specifically, we focused on the wake behind a cylinder in a two-dimensional flow. 
    The specific experimental setup is detailed in SI Document S2.2. 
    We built a 1024$\times$1024 mesh for simulation, and each one of 1M grids was regarded as an element, 
    containing the position, pressure, and velocity of the grid as the data fields, and 2048 time frames were collected in total. 
    We collected 9 to 256 adjacent or random grids and 128 time frames as a data snippet for training.

\begin{figure}[p!]
        \centering
        \includegraphics[width=\linewidth]{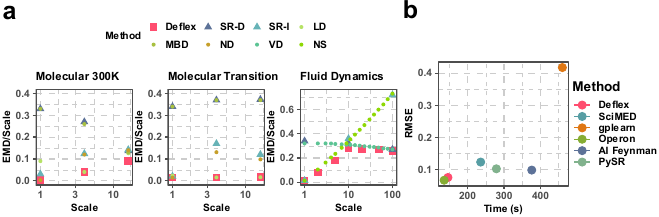}
        \caption{
            Experimental results of Deflex in terms of cross-scale performance (\textbf{a}) and benchmark performance (\textbf{b}).
            \textbf{a} shows the EMD values of known formulas
            and formulas discovered by Deflex and other methods across different coarse-graining scales 
            for water particles' motion and fluid dynamics, 
        where ``SR-I'' and ``SR-D'' represent the symbolic regression methods with invariant and distributional form settings, respectively.
        \textbf{b} shows the running time and relative mean square error for Deflex and other methods on the public benchmark dataset of Feynman Symbolic Regression Database.
    }\label{fig:analysis_cd}
\end{figure}

    To test the generalizability of our approach across different flow regimes and spatial dimensions, we conducted a second fluid dynamics experiment using the Johns Hopkins Turbulence Database (JHTDB)~\cite{li2008public}.
    Unlike the confined cylinder wake flow, the JHTDB provides Direct Numerical Simulation of three-dimensional forced isotropic turbulence, representing a canonical case of fully developed turbulence without boundary effects.
    The dataset features a $1024^3$ computational grid with a Taylor microscale Reynolds number $Re_\lambda \approx 433$.
    We extracted volumetric subsamples from the database, with each grid point containing the three-dimensional velocity field $(u, v, w)$, pressure $p$, and their spatial derivatives.
    Following the same training protocol as the cylinder experiment, we sampled 9 to 256 spatial locations and 128 temporal snapshots as input data.
    Remarkably, Deflex successfully identified the three-dimensional Navier-Stokes equation from this turbulent dataset (Table \ref{tab1}, Figs. \ref{fig:analysis_ab} and \ref{fig:analysis_cd}), capturing the full three-dimensional convective and diffusive terms.
    The complementary nature of these two experiments—2D laminar wake versus 3D turbulent flow—demonstrates Deflex's capability to discover governing equations across different flow regimes, spatial dimensions, and Reynolds numbers.

    As listed in Table \ref{tab1}, Deflex detects the Navier-Stokes equation of fluid dynamics through small-scale observations, which cannot be discovered by other methods. 
    In larger-scale random observations, Deflex extracts the velocity distribution of the fluid within the vortex flow. 
    These formulas correspond to different manifestations of long-standing laws in physics. For complex distribution formulas like velocity distribution, most compared methods are infeasible (shown in Table 2), and only AI Feynman and Deflex manage to discover.
    Notably, from the discovered Navier-Stokes equation, we can further identify the term corresponding to Newton's viscosity law, which is also listed among our discoveries in Tables \ref{tab1} and 2.
    Meanwhile, SINDy and Operon can recover a simplified form of 2D Navier--Stokes without pressure-field terms (i.e., $\nabla p$), which leads to noticeably larger errors than Deflex, even though the simpler structure can make their discovery faster. Deflex is the only method that can recover the full 3D Navier--Stokes equation in the JHTDB setting.

    Fig.~\ref{fig:analysis_ab}a also shows that the equation-based formulas discovered by Deflex can achieve a similar RMSE as other methods. For the velocity distribution, Deflex's discovery has a much smaller NLL loss than AI Feynman's, showing its better accuracy.
    Fig.~\ref{fig:analysis_ab}b,
    most comparatives, except for SINDy and Operon, become much slower than Deflex with the increase of formula complexity. Among them, AI Feynman is around 10 times slower than Deflex.
    
    In addition, we compared the EMD across different coarse-graining scales of Deflex with that of AI Feynman and the renormalization of known formulas.
    Fig. \ref{fig:analysis_cd}a shows that predictions from Deflex keep the lowest EMD across different scales,
    indicating the Deflex's ability to overcome the scale gap issue in complex systems.
    Meanwhile, other methods with a single setting can only capture the patterns at a single scale, and the EMD increases significantly when the scale changes.
    
    \subsection{Comparison on benchmark dataset}
    We finally evaluated Deflex on the public benchmark dataset of Feynman Symbolic Regression Database~\cite{doi:10.1126/sciadv.aay2631}. This benchmark dataset is generated by sampling from 120 classical formulas spanning different fields of physics, with added noise to simulate real-world conditions. It captures essential characteristics of real-world mathematical formulas, such as symmetry, while also being a classic, well-established set of problems for symbolic regression. This dual characteristic facilitates a feasible performance comparison against other SR-based AI methods.
    In the experiment, Deflex was compared against Operon, SciMED, gplearn, AI Feynman 2.0, and PySR. We compared their performance by measuring the trade-off between the average RMSE of discovered formulas and the total runtime for the discovery.
    Note that SINDy was not included as it is sparse regression-based and not designed for symbolic regression tasks, while DEAP was omitted since it works as the backend of gplearn in the current task.
    
    As shown in Fig. \ref{fig:analysis_cd}b, Deflex demonstrates quite competitive performance among all compared methods on the benchmark dataset in terms of the accuracy-efficiency tradeoff, although it is not specialized for these traditional, closed-form equations. 
    In particular, Deflex is only slightly outperformed by Operon, which relies on a fixed symbol set and is optimized for runtime. 
    In our experiment, Deflex managed to discover the formula with an RMSE of 0.075 in just 146.2 seconds, which was quite close to Operon with an RMSE of 0.0663 in 135.8 seconds. 
    And Deflex and Operon show significant advantages over other methods.
    This high efficiency is largely attributed to the pre-training mechanism, which accelerates formula discovery by incorporating mathematical priors (Fig.~\ref{fig:ablation}a, d).

    \subsection{Ablation analysis}
\begin{figure}[h]
        \centering
        \includegraphics[width=1.0\linewidth]{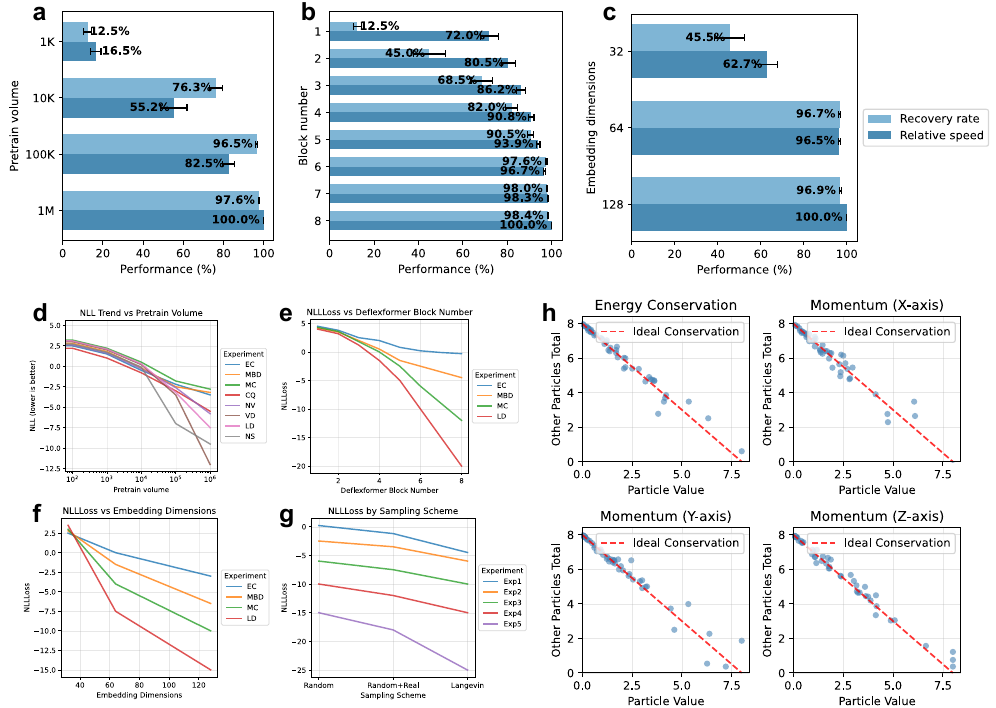}
    \caption{Ablation studies and component analysis of Deflex. \textbf{a}, \textbf{d}, Impact of pre-training volume on formula discovery performance (recovery rate and relative speed) and NLL trend. Larger pre-training volumes significantly accelerate discovery and improve accuracy. \textbf{b}, \textbf{e}, Performance metrics and NLL loss with varying Deflexformer block numbers. Increasing depth improves representation power. \textbf{c}, \textbf{f}, Effect of embedding dimensions on performance. \textbf{g}, NLL loss comparison across different sampling schemes (Random, Random+Real, Langevin). Langevin sampling yields the best fitness. \textbf{h}, Verification of conservation laws (Energy and Momentum) learned by the Deflexformer, showing close alignment between sampled states (black dots) and ideal laws (red lines). In panels \textbf{a}--\textbf{c}, the error bars show mean value $\pm$ standard deviation.
    }
        \label{fig:ablation}
    \end{figure}

    To validate the contribution of each component in Deflex, 
    we conducted systematic ablation studies across multiple dimensions (Fig.~\ref{fig:ablation}).
    We first examined the impact of pre-training on discovery efficiency by varying the synthetic dataset size from 0 to $10^6$ formulas.
    Results demonstrate that larger pre-training volumes lead to substantially faster convergence and higher formula recovery rates (Fig.~\ref{fig:ablation}a, d).
    Pre-training with $10^6$ examples achieved over 80\% recovery rate within 5,000 steps, 
    while training from scratch required more than 20,000 steps to reach similar performance,
    confirming that incorporating mathematical priors through pre-training is crucial for efficient exploration of the formula space.
    
    Architectural ablations reveal that both network depth and embedding dimensions critically affect the model's capacity to capture complex system dynamics.
    Increasing the number of Deflexformer blocks from 1 to 8 progressively improves performance, 
    with NLL loss decreasing from approximately 1.8 to 0.6 (Fig.~\ref{fig:ablation}b, e).
    Similarly, expanding embedding dimensions from 32 to 128 enhances the model's representational power (Fig.~\ref{fig:ablation}c, f).
    However, performance saturates beyond 6 blocks and 128 dimensions, 
    indicating that these configurations provide sufficient capacity for the tested systems.
    We also conducted ablation studies on the derivative (gradient) features from the datasets, as detailed in the Supplementary Information, finding that accuracy is mostly preserved but discovery efficiency degrades, and the complicated 3D Navier--Stokes case (JHTDB) does not reliably converge under the same time budget.
    Moreover, the Deflexformer effectively internalizes physical constraints, 
    autonomously learning energy and momentum conservation manifolds (Fig.~\ref{fig:ablation}h).
    Finally, comparing data sampling strategies demonstrates that Langevin sampling 
    is essential for accurately identifying complex equilibrium distributions (Fig.~\ref{fig:ablation}g).
    
    \section{Discussion}\label{sec13}
    
    We present an AI-based framework Deflex for scientific rules discovery from complex systems, which can extract multiscale mathematical formulas through neural-network-guided decomposition and symbolic regression. 
    In Deflex, we propose a unified energy model and distribution-based representation of formulas, which enables multiscale probability modeling, 
    while leveraging lambda calculus extends symbolic regression's capacity to handle massively parallel entity interactions without prior structural assumptions.
    Deflex first trains a neural network to decompose intricate system interactions into elementary components, 
    then interprets these into compact mathematical formulas using lambda-calculus-based symbolic regression.
    By breaking down formula learning into tractable sub-problems, this decomposition strategy solves the issue of exponential complexity in traditional symbolic regression. 
    
    Extensive experiments on six typical complex systems across diverse domains, ranging from particle dynamics and fluid mechanics to human and bird mobility, demonstrate the effectiveness and efficiency of Deflex in multiscale formula discovery. Experimental results demonstrate that Deflex can manage to overcome the scale gap issue by achieving the least differences across scales. Compared with state-of-the-art single-scale methods, it can also achieve better performance in both accuracy and computational efficiency in most cases, where the runtime reduction can achieve up to orders of magnitude. In addition to reconstructing established principles (e.g., Langevin equations, energy conservation), Deflex also shows the potential to reveal undocumented system relationships from the observed data.
    
    The critical advancement of Deflex lies in the multiscale characterization. 
    As evidenced by the experiments on ice crystals and fluid dynamics, 
    Deflex can capture scale-consistent patterns without renormalization procedures, surpassing traditional multiscale description methods. 
    This advantage stems from the synergy between the invariant modeling of EBMs and the compositional flexibility of lambda calculus,
    eliminating the prerequisite for explicit scale separation or prior formula knowledge. 
    In contrast, conventional methods require substantial prior specification of the potential formula structures, 
    severely limiting their autonomous discovery capability. Nonetheless, on the benchmark AI Feynman dataset, we observe that Deflex has a lower recovery rate on formulas of quantum transition probability derived from time-dependent perturbation theory~\cite{sakurai2017modern} 
    (see SI for detailed introduction and the recovery comparison). 
    We guess that this phenomenon exposes a potential limitation of Deflex's multiscale discovery. That is, when candidate structures of markedly different magnitudes coexist, low-frequency approximations may overwhelm the original high-frequency but complex structures. This suggests an open problem in balancing competing scales so that small yet physically meaningful terms are not suppressed.
    
    From a practical application perspective, Deflex requires that the observation of complex systems can be organized into learning samples. In a typical physical setting, each sample combines global context with evolving states of a set of $n$ interacting entities over consecutive time steps. In practice, one can obtain a long continuous record and then construct training samples by extracting spatiotemporal sub-trajectories, allowing the model to learn an energy-form expression that aggregates variable-size interactions. Successful symbolic recovery further demands sufficiently diverse and broad coverage of the system's state space. This can be done by collecting multiple trajectories under varied initial conditions. If data collection is limited, sampling over multiple windows and entity subsets may help to increase the number of training samples without altering the underlying physics. In real applications, the scientist-in-the-loop approach can also be incorporated into data collection and feature engineering when domain knowledge is needed.

    While demonstrating significant progress, several limitations in the current implementations still warrant further investigation. 
    First, the computational efficiency of deep formula synthesis remains constrained by excessive invalid candidate generation in symbolic regression.
    Second, the current framework assumes pre-processed observational data, whereas real-world applications may require integrated solutions for automated data acquisition and hypothesis generation. Therefore, future work may explore improving symbolic search efficiency via hybrid inference strategies combining neural guidance with type-theoretic constraints. Third, regarding interferences of the approximated patterns among multiple scales, a promising direction is to separate scales explicitly in the discovery procedures, including problem formulation, rather than extracting implicitly within a single joint run.
    Also, it is important to establish theoretical connections between the learned energy landscapes and renormalization group theory. Another practical problem is handling noisy, incomplete observational data to enhance the robustness for real-world complex systems.
    
    In this work, we have presented Deflex, a framework for discovering governing equations from multiscale complex systems. 
    Our empirical results demonstrate its ability to identify symbolic rules across different scales and, in some cases, 
    reveal previously undocumented relationships from observational data. 
    The proposed methodology for representing interactions among high-dimensional entities suggests potential applicability to domains beyond those tested in our experiments. 
    Future work could explore the practical meaning of the undocumented formulas with area experts, and Deflex's further use in areas such as atmospheric dynamics modeling~\cite{bi2023accurate} and materials science~\cite{benzi1984multifractal}, 
    particularly for systems where strong scale interdependencies pose challenges to conventional methods. We hope that this work serves as a step toward more capable automated scientific discovery in a range of disciplines.
    
    \section{Methods}
    
    We present the details of the Deflex framework for discovering mathematical laws in complex systems, which can automatically learn concise mathematical formulas from the observed data of the systems. 
    
    Throughout this section, 
    we use bold symbols to denote vectors and matrices (e.g., $\mathbf{x}$ for individual samples, $\boldsymbol{\theta}$ for model parameters). 
    Specifically, 
    $\mathbf{X}=\{\mathbf{x}_1,\ldots,\mathbf{x}_N\}$ represents the observation dataset with $N$ samples, 
    each $\mathbf{x}_i$ characterizing the dynamics of element $i$ over time.
    We denote theoretical energy functions as $E(\cdot)$ or $E^*(\cdot)$ for the optimal one,
    and use $\mathcal{E}_{\boldsymbol{\theta}}(\cdot)$ to refer to the parameterized energy function approximated by the Deflexformer neural network.
    The partition function is denoted as $Z_{\boldsymbol{\theta}}$, 
    and probability distributions as $p(\cdot)$ for data distributions and $Q_{\boldsymbol{\theta}}(\cdot)$ for model distributions.
    
    \subsection{Problem formulation}

    We consider describing a complex system by a set of measurable features $\mathbf{x}$, which normally include the element-level ones $\mathbf{s}$ and optionally the global-level ones $\mathbf{v}$, i.e., $\mathbf{x}=[\mathbf{s};\mathbf{v}]$.
    Specifically, the element-level features $\mathbf{s}$ are state variables that describe the local, 
    dynamic condition of individual components (or elements) within the system (e.g., a particle's position and velocity), which are time-varying. The global-level features $\mathbf{v}$ may characterize the system from a macroscopic level (e.g., a system-wide temperature) and is considered independent of both element amount and time.
    In particular, 
    Suppose we can obtain an observation dataset  $\mathbf{X}=[\mathbf{S}; \mathbf{V}]$ from the system, where $\mathbf{S}\in \mathbb{R}^{NTD_s}$ denotes the $D_s$-dimensional state $\mathbf{s}$ of $N$ elements within a time period of $T$, and the optional $\mathbf{V}\in \mathbb{R}^{D_v}$ represents the $D_v$ dimensional global information $\mathbf{v}$.
    Note that, except for the category of global or elementary features, we have no prior knowledge about the relationships among the individual features.

    Let $\mathcal{F}$ denote the set of all possible mathematical formulas, including equations, invariants, and distributions.
    The behaviors in the system can be described by a set of formulas $\mathbf{f} \subseteq \mathcal{F}$ on $\mathbf{X}$.
    Normally, $\mathbf{f}=\{f_1,\cdots,f_m\}$ (where $\forall f_i \in \mathbf{f}, f_i \in \mathcal{F}$) with $m$ formulas is expected to satisfy two properties:
    1) $\mathbf{f}$ should fit the observed data well, i.e., have a large likelihood. Specifically, we can define a certain loss function $\mathcal{L}(f_i, \mathbf{X})$ to measure how well $f_i \in \mathcal{F}$ fits the datasets $\mathbf{X}$, where lower loss indicates better fit.
    2) $\mathbf{f}$ should be concise, that is, have low complexity in mathematical formulas, denoted as $R(f)$.
    Then, the goal of Deflex is to find the mathematical formulas $\mathbf{f^*} \subseteq \mathcal{F}$ that balance the trade-off 
    between the complexity of the expression of the function and the fitness of the data (in terms of the expected loss over the observed data $\mathbf{X}$).
    This can be formulated as the Pareto optimal set of patterns $\mathbf{f^*} \subseteq \mathcal{F}$ that satisfies:
    \begin{equation}
        \label{eq:opt}
        \forall \mathbf{g} \subseteq \mathcal{F}, \ {\sum\nolimits_{f \in \mathbf{f^*}} \mathcal{L}(f, \mathbf{X})}  \leq {\sum\nolimits_{g \in \mathbf{g}} \mathcal{L}(g, \mathbf{X})}  \Rightarrow {\sum\nolimits_{f \in \mathbf{f}}R(f)}\leq {\sum\nolimits_{g \in \mathbf{g^*}}R(g)}.
    \end{equation}

    This is essentially a symbolic regression (SR) problem.
    Existing SR methods explore to solve the above problem by evaluating large-scale candidate expressions. However, facing complex systems with complicated observation data and multiscale rules, existing studies would suffer from three major challenges:
    \begin{itemize}
    \item Limited expressibility for higher-order relationships: traditional expression trees struggle to represent mappings and higher-order functions autonomously, often resorting to rigid, pre-specified operators or suffering from explosive tree growth when modeling collective behaviors.
    \item Disjoint representations across scales: existing approaches lack a unified framework to characterize and learn both deterministic laws (invariants) and stochastic patterns (distributions) that emerge at different scales.
    \item Inefficiency in complex search spaces: the search space for complex nonlinear formulas grows exponentially with the number of variables, making standard search algorithms computationally prohibitive for high-dimensional systems.
    
    \end{itemize}
    
    \subsection{Distribution-based formula representation}
    We first unify formulas in the form of probability distributions, which can be represented as energy functions that map the observation data $\mathbf{X}$ (with both element-level state information $\mathbf{S}$ and the optional global context $\mathbf{V}$) to scalar-valued energy $f_1(\mathbf{X}), \ldots, f_m(\mathbf{X})$. Then the search space $\mathcal{F}$ of Deflex is unified as the set of energy functions.

    \subsubsection{Unified formula representation with distributions.}\label{method:uni}
    For a unified representation, 
    we can regard any formula $f\in \mathcal{F}$ as a probability distribution $p(\mathbf{x})$~\cite{siegenfeld2020introduction}.
    Specifically, 
    a collective formula $\mathbf{f}=\{f_1,\cdots,f_m\}$, 
    describing the behaviors of a single system, 
    can be seen as a set $\mathbf{p}=\{p_1,\cdots,p_m\}$ of $m$ independent distributions $p_i(\mathbf{x})$ over $\mathbf{x}$, 
    i.e., $\mathbf{p} \equiv \mathbf{f}$.
    Among them, 
    equations and invariants are treated as distributions with very narrow peaks,
    which indicates their tiny errors or uncertainty.
    The original distribution-based formulas are treated as distributions with wider peaks. 
    This representation unifies different forms of formula expressions, 
    thus simplifying the formula discovery.
    Furthermore, 
    any two independent distributions $p_i(\mathbf{x}), p_j(\mathbf{x}) \in \mathbf{p}$ can be equivalently described as a single mixed distribution $p'_{\{i,j\}}(\mathbf{x})$ of them.
    Ultimately, 
    $\mathbf{p}$ with $m$ independent distributions can also be seen as a set $\mathbf{p}'=\{p'_{\{1,\ldots, m\}}\}$ with a single mixed distribution $p'_{\{1,\ldots, m\}}(\mathbf{x})$ of $p_1(\mathbf{x}),\cdots, p_m(\mathbf{x})$.
    Then, 
    we also say $\mathbf{p}'=\mathbf{p} \equiv \mathbf{f}$. 
    This aligns well with the characteristics of different scales in complex systems and naturally resolves the scale gap issue in multiscale pattern discovery.
    That is, 
    probability distributions on a large scale can describe the behaviors on a small scale. 
    In other words, 
    narrow distributions correspond to precise small-scale patterns,
    whereas wider distributions correspond to large-scale patterns.
    Based on the unified representation, 
    solving the problem in Eq.~(\ref{eq:opt}) requires discovering an optimal distribution $p^*(\mathbf{x})$ that best fits (with the maximum likelihood of) $p_\text{data}(\mathbf{x})$, 
    which denotes the underlying data distribution of the observation dataset.

    \subsubsection{Characterizing distributions with energy functions.}\label{method:ebm}
    The unified representation can tackle the scale gap issue in multiscale systems.
    However, 
    without the ability to normalize probability density functions, 
    SR cannot directly discover the distribution-based formulas.
    Therefore, 
    we further characterize distributions as energy functions under the energy-based models (EBM~\cite{song2021train,wenliang2019learning,song2019generative, roney2022state}) framework, 
    which are less restrictive in functional forms (i.e., without requiring a normalized probability).
    
    EBMs are probabilistic models inspired by the Boltzmann distribution in statistical physics.
    Let $p(\mathbf{x})$ denote a distribution over features $\mathbf{x}$.
    Then, 
    $p(\mathbf{x})$ can be characterized as an energy function $E(\mathbf{x})$, 
    which indicates different energy values over the features $\mathbf{x}$.
    A probability distribution function (PDF) $p(\mathbf{x})$ given by an EBM is:
    \begin{equation}
        \label{eq:ebm}
        p(\mathbf{x})= \frac{1}{Z}\exp(-E(\mathbf{x}))
    \end{equation}
    where $Z= \int \exp(-E(\mathbf{x})) d\mathbf{x}$ is the normalization factor (i.e., partition function) and ensures the integral of the distribution equals 1.
    It can avoid complex normalization calculations and volume transformations for complex and multi-dimensional variables. 
    Therefore, 
    finding the optimal distribution $p^*(\mathbf{x})$ is essentially equivalent to finding the optimal energy function $E^*(\mathbf{x})$.
    In our framework,
    we use the Deflexformer neural network $\mathcal{E}_{\boldsymbol{\theta}}(\mathbf{x})$ to approximate $E^*(\mathbf{x})$,
    which can be achieved by training on observation data.

    \subsection{Deflexpressor: SR subsystem based on lambda calculus}\label{method:expr}
    Traditional tree-based SR relies on first-order expression trees and manually specified higher-order primitives, which makes it difficult to learn relationships such as summation or mapping over massive interacting elements in complex systems. To this end, we present Deflexpressor, an SR system based on the extended lambda calculus, which supports various higher-order expressions while ensuring valid types and variable length of elementary inputs. In particular, Deflexpressor leverages a strict rule-based generator together with an efficient neural network-based one to generate a large number of valid candidate lambda expressions for mathematical relationships, which are then evaluated to identify Pareto-optimal expressions in terms of accuracy and conciseness. 
    
    \subsubsection{Expressing relationships by lambda calculus}
    
    Lambda calculus is a formal system in mathematical logic and computer science for expressing computation based on function abstraction and application~\cite{barendregt1984lambda}. A lambda term is built from three kinds of expressions: (i) Variable $x$ denotes a placeholder for a value or parameter; (ii) Abstraction $\lambda x. M$ defines a function with parameter $x$ and body $M$; (iii) Application $M N$ applies function $M$ to argument $N$. Although there is no built-in data, the lambda calculus can encode data and control structures (e.g., recursion) as functions via Church encoding. The lambda calculus is Turing-complete and can simulate any computation based on symbol replacement. Specifically, for any expression made of variables, symbol $\lambda$, and parentheses, we can find a pattern $(\lambda x. M) N$ and replace it with $M$ where every $x$ is replaced by $N$. By repeating this process until no such pattern remains, we can obtain the final expression as the result. The basic grammar rules are provided in Section 3.1 in SI.
    
    Since the lambda calculus is Turing-complete and natively supports function abstraction and recursion, Deflexpressor can express operations such as summation and mapping directly rather than as fixed templates. Specifically, operations over collections can be written as recursively defined functions or higher-order maps, enabling the system to learn how to aggregate interactions from data without requiring pre-specified structures. 
    For instance, we consider a common formula $E = \sum_{i \neq j} V(r_{ij})$ that describes the aggregation of a variable number of particle interactions. 
    Here, $V(r_{ij})$ denotes a function that calculates the pairwise interaction potential based on the distance $r_{ij}$ between two particles indexed by $i$ and $j$. Traditional SR either requires hard-coding symbols or extremely deep trees to express such a relationship.
    In contrast, suppose that $\mathbf{x}=[r_{ij}]_{i\neq j}$ denotes the list that enumerates these distances under any fixed ordering, Deflexpressor can naturally write the aggregation as the following Lambda expression.
    
    \begin{equation}
    \begin{aligned}
    \underbrace{
    \Big(
    \underbrace{\mathrm{Fix}\Big(
    \underbrace{
    \underbrace{\lambda f\,\lambda \mathrm{xs}\,\lambda a.\rule[-4.7ex]{0pt}{0pt}}_{\text{Variables}}
    \underbrace{\underbrace{
    \mathrm{if}\;\mathrm{Null}(\mathrm{xs})\;\mathrm{then}\;a}_{\text{Application (i)}}\;\mathrm{else}\;
    \underbrace{f\;(\mathrm{Tail}(\mathrm{xs}))\;\big(a + V(\mathrm{Head}(\mathrm{xs}))\big)}_{\text{Application (ii)}}
    }_{\text{Function body describing the whole relationship (i and ii) among $f,\mathrm{xs}, a$}}
    }_{\text{Abstraction defining a function G(f, xs, a) over variables $f,\mathrm{xs}, a$}}
    \Big)}_{\text{Fixed-point simulating a recursive function F(xs, a) based on G(f, xs, a)}}
    \Big)
    \;\mathbf{x}\;0\,}_{\text{Application denoting applying recursive function F(xs, a) over parameters $\mathbf{x}$ and $0$}}.
    \end{aligned}
    \label{eq:fix_sum}
    \end{equation}

    Eq.~(\ref{eq:fix_sum}) defines a recursive aggregation function $F$ via a fixed-point construction and applies it to a variable-length list $\mathbf{x}$ and an initial accumulator $0$.
    The underbraces decompose the term: the innermost abstraction defines a one-step update rule $G(f,\mathrm{xs},a)$ whose body implements (i) a base case that returns $a$ when $\mathrm{Null}(\mathrm{xs})$ and (ii) an update-and-continue step that calls $f$ on $\mathrm{Tail}(\mathrm{xs})$ with the updated accumulator $a+V(\mathrm{Head}(\mathrm{xs}))$.
    The operator $\mathrm{Fix}(\cdot)$ turns this update rule into a self-referential recursion $F$ (formal definition in the SI), and application associates to the left, so $F\ \mathbf{x}\ 0$ means $(F\ \mathbf{x})\ 0$.
    See SI for a step-by-step expansion and an explicit walk-through of the bracketed decomposition.
    This demonstrates that the system constructs the algorithm of summation from scratch, 
    covering cases where predefined summation templates are inapplicable or insufficient.
    
    Besides strong expressiveness, another merit of the lambda calculus is that its relatively unified structure enables intuitive comparison of the complexity of formulas. In particular, we can directly use the expression length as the complexity metric $R(f)$ (in Eq.~\ref{eq:opt}) in Pareto selection, enabling an explicit trade-off between data fitness and conciseness.

    \subsubsection{Extended lambda calculus with types and arrays}
    The pure lambda calculus is Turing complete, but it has two limitations in our complex system setting. The first is that it may generate meaningless expressions and even non-terminating reductions, which would lead to the low efficiency of the symbolic regression process in Deflexpressor. The second is that data structures like lists or arrays are often encoded as nested higher-order functions, which leads to significantly high complexity in data traversal. To this end, in Deflexpressor, we adopt an extended lambda calculus with types and arrays.
    
    A type specifies what kind of value a sub-expression denotes (e.g., a real number, a vector, or a function), and therefore constrains how expressions can be composed. Specifically, we enforce the following type constraints.
    (i) Type compliance in the applications. For example, an application $(M\,N)$ is valid only when $M$ denotes a function whose input type matches the type of $N$. Instead, if $M:\mathbb{R}$ denotes a real-valued number, then $(M\,N)$ will be rejected.
    (ii) Type requirements for the operands of operators.
    For instance, expression $M = 1 + (\lambda x.\,x)$ will be identified as invalid because $\lambda x.\,x$ (i.e., $f(x)=x$) belongs to a function type, whereas $+$ expects two operands with the real-valued type.
    Type constraints on input-output format.
    For example, many tasks require a real-valued function on an $n$-dimensional state, i.e., $f:\mathbb{R}^n\rightarrow\mathbb{R}$. In such cases, a well-typed function term can be used, whereas a constant will be rejected in Deflexpressor.
    For instance, $E=\sum_{i\neq j}V(r_{ij})$ requires a mapping from a variable-length list of pair distances $\mathbf{x}=[r_{ij}]_{i\neq j}$ to a scalar energy.
    Eq.~(\ref{eq:fix_sum}) has exactly this form (a function that consumes $\mathbf{x}$ and returns $E$), thus being a type-valid expression.
    In Deflexpressor, we implement a type inference engine based on the classical type inference algorithm W~\cite{MILNER1978348} (see SI for details), which can automatically infer the type of each candidate expression, and then keep only type-valid candidates.
    
    Variable-sized data structures, such as lists, arrays, and vectors, are essential to represent various data in complex systems. However, encoding such structures as functions in pure lambda calculus would lead to extra workload of searching the encoding itself rather than the relationship of encoded data. To adapt the framework for scientific discovery, we predefine array $\text{Array}[\tau]$ as a built-in primitive data type, and provide $O(1)$-time access and basic operators (e.g., indexing and size). For instance, we treat the pair distances $r_{ij}$ in $E=\sum_{i\neq j}V(r_{ij})$ as an array $\mathbf{x}=[r_{ij}]_{i\neq j}$. This allows the aggregation to be written in the same form as Eq.~(\ref{eq:fix_sum}), letting the search focus on learning $V(\cdot)$ and the aggregation structure rather than on data representation details (formal definitions in SI).

    \subsubsection{Hybrid expression generation with rules and NNs}
    Exploring the vast space of lambda expressions presents distinct challenges in terms of efficiency and coverage.
    Firstly, rule-based mutations (e.g., random subtree replacement) explore the space diffusively and often generate low-quality expressions. To boost search efficiency, we incorporate a causal autoregressive neural network based on the Transformer Decoder architecture (see SI for architecture details).
    This network is trained offline via self-supervised learning on a manually-prepared corpus of valid expression trees and their rule-based mutation pairs.
    By learning the probabilistic priors of valid structural transformations,
    the network can generate syntactically correct and type-consistent lambda expressions either from scratch or by mutating existing ones. Note that, the produced expressions are parsed and type-checked to ensure validity.
    For example, a subtree mutation can replace the local update term $V(\mathrm{Head}(\mathrm{xs}))$ in Eq.~(\ref{eq:fix_sum}) with $V'(\mathrm{Head}(\mathrm{xs}))$ without changing the recursion structure.
    $\beta$-reduction further simplifies candidates by evaluating function applications, e.g., $((\lambda x.\,x)\ y)\rightarrow y$ (formal rules in SI).
    Secondly, relying solely on a neural network can lead to mode collapse, where the search gets stuck in familiar patterns and misses novel solutions. To expand search coverage, we implement a hybrid sampling strategy with simulated annealing. The process transitions from predominantly rule-based generation (for broad exploration) to neural-guided generation (for exploitation).
    Third, expressions after multiple mutations often become bloated with redundancy.
    To maintain compact and interpretable forms, we enforce lambda-calculus-specific mutations like $\beta$-reduction and common sub-expression elimination in the rule-based component (see SI for mutation details). 
    
    \subsubsection{Initialization with global aggregation priors}
    While Deflexpressor can generate expressions of any computable functions via the lambda calculus,
    standard random initialization tends to produce expression trees that operate on a fixed number of arguments.
    In such a case, it would be combinatorially expensive to evolve a structure that iteratively computes over a variable-length input from scratch.
    
    To accelerate this process without imposing hard constraints, we employ a heuristic initialization strategy that generates the seed expressions performing various aggregations over variable-length inputs.
    Specifically, these seed expressions are written in the form $\lambda \mathbf{x}.\ \text{Reduce}(\mathrm{kernel}, \mathbf{x})$, which denotes the recursive procedure of updating an aggregation state $s$ by traversing a variable-length input list $\mathbf{x}=[x_1,\dots,x_N]$ as follows
    \begin{equation}
        \label{eq:reduce}
        s \gets \left\{\begin{array}{l}
        x_1, i=1 \\
        \text {kernel}\left(s, x_i\right), i>1
    \end{array}\right.
    \end{equation}
    where $\mathrm{kernel}(s,x_i)$ is a learnable binary operator that updates the aggregation state $s$ using an element $x_i$ of the list $\mathbf{x}$.
    
    In particular, the $\mathrm{kernel}$ is a randomly generated expression tree that defines the interaction logic between the aggregation state $s$ and each data element  $x_i$ of a variable-length list.
    For example, in $E=\sum_{i\neq j}V(r_{ij})$, we can define a kernel as $\mathrm{kernel}(s,r)=s+V(r)$ and take $\mathbf{x}=[r_{ij}]_{i\neq j}$ as the list.
    Then the function $\text{Reduce} (\mathrm{kernel},\mathbf{x})$ can yield the desired aggregation, i.e., it stands for computing $E$ by iteratively applying $s\leftarrow s+V(r)$ over all pairwise distances $r_{ij}$ in $\mathbf{x}$ (with the recursive procedures realized via $\mathrm{Fix}(\cdot)$ as in Eq.~(\ref{eq:fix_sum})).
    This initialization serves as a soft inductive bias that prioritizes aggregation operations over massive elements, encouraging SR to focus immediately on discovering local interaction laws (encoded in the $\mathrm{kernel}$) while fitting the data.
    Meanwhile, since $\lambda \mathbf{x}.\ \text{Reduce}(\mathrm{kernel}, \mathbf{x})$ is also a recursive lambda expression that can be implemented with the same recursion primitive $\mathrm{Fix}(\cdot)$ in Eq.~(\ref{eq:fix_sum}), it remains fully mutable under subsequent lambda calculus mutations.
    Subsequent evolutionary steps (e.g., subtree mutation, $\beta$-reduction) can alter the $\mathrm{kernel}$, break the recursion, or discard this ``Reduce-like'' structure entirely if a simpler non-recursive law fits the data well.

    \subsection{Deflexformer: energy-based NNs with decomposable blocks}\label{method:former}
    The optimal energy function $E^*(\cdot)$ can theoretically be inferred via SR over the observation dataset.
    However, this can be computationally expensive and even infeasible for complex expressions.
    Generally, any complex $E(\cdot)$ can be expressed as the composition of a series of (e.g., $k$) base functions $E^i(\cdot)$, e.g., $E(\cdot)=E^1 \circ E^2\circ \ldots \circ E^k(\cdot)$.
    So, we propose to approximate $E(\cdot)$ by training a deep neural network Deflexformer $\mathcal{E}_{\boldsymbol{\theta}}(\cdot)$ (parameterized by $\boldsymbol{\theta}$) with $k$ decomposable blocks $\mathcal{E}^i(\cdot)$, each approximating a base function $E^i(\cdot)$ (where $i=1,\ldots,k$).
    The decomposable structure enables penetrating each block $\mathcal{E}^i(\cdot)$ to derive explicit expressions of $E^i(\cdot)$, making SR feasible to discover complex expressions of $E(\cdot)$.

    \subsubsection{Architecture of Deflexformer}
    The Deflexformer network predicts the energy value for a given dataset using the aligned state sequences of elements and global context as inputs. 
    The architecture comprises an input embedding layer, $k$ stacked Deflexformer blocks, and an output projection module.
    
    The input layer embeds raw observations $\mathbf{X}$ into a high-dimensional feature space via Fourier multiscale embedding (see SI), producing element-level representations $P_0 \in \mathbb{R}^{n \times t \times D_{model}}$ and global representations $G_0 \in \mathbb{R}^{c \times D_{model}}$, where $n$ is the number of elements, $t$ is the number of time frames, $c$ is the number of global tokens, and $D_{model}=64$ is the model dimension. For some observations of a complex system, $\mathbf{X}$ may only include the element-level state information $\mathbf{S}$ without the global information $\mathbf{v}$. In such a case, we set a dummy global feature $\mathbf{v}$ and randomly generate a dummy global context dataset $\mathbf{V}$ by drawing some random noise over $\mathbf{v}$ (see SI for details).
    Note that, designed for a much more complicated SR problem over observation data of complex systems with multiple scales, Deflexformer can also be directly used for classical SR tasks (e.g., applied to the AI Feynman dataset in Section~2.7), which correspond to a single scale without considering element-level dynamics. Here, the SR dataset is input as the global context, and the element-level representation is left empty at the input layer.

\begin{figure}[p!]
        \centering
        \includegraphics[width=0.7\linewidth]{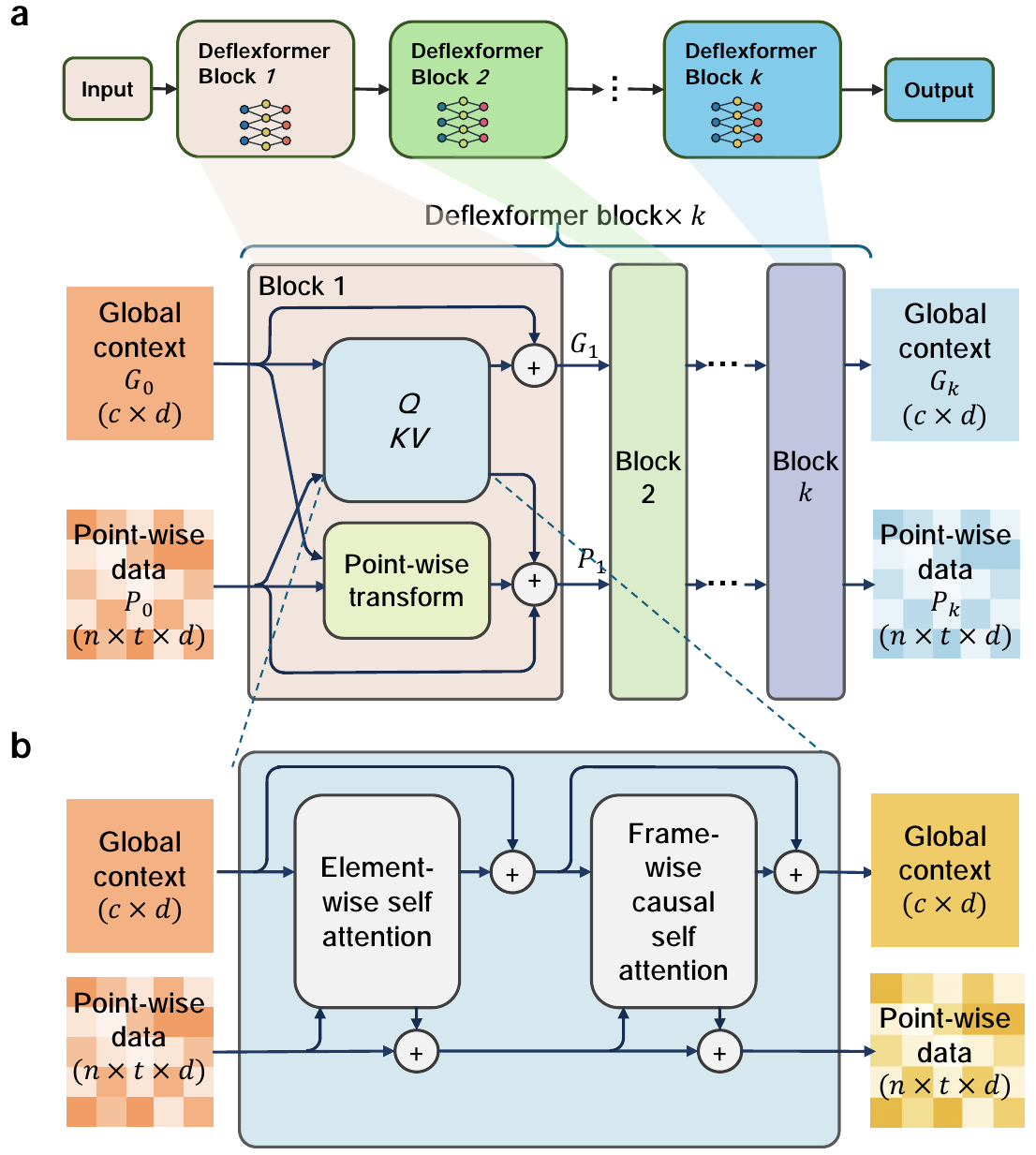}
        \caption{
            Illustration of the neural network architecture of Deflex.
            \textbf{a} illustrates the overall model architecture of Deflexformer.
            Arrows show the information flow among the various components.
            The intermediate expressions are separated into element-wise and global variables
            and are processed by different neural networks.
            \textbf{b} illustrates the self-attention network for a Deflexformer block.
            The point-wise data consists of $t$ time frames,
            each containing $n$ elements with $d$ dimensions.
            The point-wise data is first processed by the element-wise residual self-attention with the fusion of global variables
            and then processed by the frame-wise residual causal self-attention with the fusion of global variables.
        }\label{fig:arch}
    \end{figure}

    Each Deflexformer block $\mathcal{E}^i(\cdot)$ (Fig. \ref{fig:arch}a) transforms representations $(P_{i-1}, G_{i-1})$ into $(P_i, G_i)$, approximating a base mathematical transformation $E^i(\cdot)$ in the energy decomposition $E(\cdot) = E^1 \circ E^2 \circ \ldots \circ E^k(\cdot)$.
    Each block comprises two sub-modules operating on the concatenated input of dimension $(c+nt) \times D_{model}$:
    (1) A point-wise transformation network implemented as a four-layer fully connected network with architecture $(D_{model}, 2D_{model}, 2D_{model}, D_{model})$ and ReLU activations, modeling element-wise evolution (e.g., the dynamics of an individual particle).
    (2) A multi-head cross-attention network with $h=8$ heads for capturing the interactions among particles.
    The attention operates in two sequential stages (Fig. \ref{fig:arch}b):
    spatial mixing performs attention over the element dimension to capture inter-element interactions, while temporal mixing applies causal attention over the time dimension to model temporal dependencies.
    By factorizing attention into spatial and temporal axes, computational complexity is reduced from $O(nt)$ to $O(n+t)$.
    Residual connections and layer normalization are applied after each sub-module (see SI for detailed specifications).
    
    The output module projects the final global representation $G_k$ back to a scalar energy via inverse Fourier embedding $e = \mathcal{E}^{-1}(G_k)$ (see SI).
    For example, when modeling particle energy $E = \frac{1}{2}m_i \mathbf{v}_i^2 + U(\mathbf{x}_i)$, early blocks may capture kinetic energy contributions while later blocks model potential interactions, with their composition yielding the total energy.
    
    \subsubsection{Pre-training the Deflexformer block}
    Directly training Deflexformer $\mathcal{E}(\cdot)$ from scratch requires a vast amount of observation data. Yet, its blocks $\mathcal{E}^i(\cdot)$ are normally simple functions, which have a much smaller search space. To this end, we adopt a common two-stage training framework of pre-training and post-training in other transformer-based deep networks. 
    In particular, a single Deflexformer block is first pre-trained with a large amount of synthetic data based on some specified relationships. Each Deflexformer block $\mathcal{E}^i(\cdot)$ represents a basic mathematical transformation, similar to some base functions (e.g., $\sin(\cdot)$) with a simple expression form. 
    Therefore, we can pre-train the Deflexformer block on a synthetic dataset sampled from some known base functions. 
    Specifically, we first call Deflexpressor to generate a pool of expressions for functions $f(\cdot)$. Then, based on these functions, we can sample numerous data pairs $(x,y)$ to synthesize a pre-training dataset $\mathcal{D}$, where $x$ is drawn from a Gaussian distribution and $y=f(x)$.
    Here, pre-training aims to fit a block model $\mathcal{E}'(\cdot)$ via supervised learning on the synthetic dataset. So the loss function is defined simply as the MSE between the predictions with labels, given the features of data pairs in the synthetic dataset. That is,  $L_{\mathcal{E}'} = \sum\nolimits_{i=1}^{|\mathcal{D}|}\text{MSE} (y, \mathcal{E}'(x))$ (see SI for training details).
    
    \subsubsection{Post-training the full Deflexformer}
    
    Then, 
    the full Deflexformer, 
    composed of multiple copies of the pre-trained block and an output module, 
    is post-trained on the observation data. 
    Note that Deflexformer $\mathcal{E}_{\boldsymbol{\theta}}(\cdot)$ is an energy-based model for approximating the optimal energy function $E(\cdot)$.
    So, 
    its PDF $Q_{\boldsymbol{\theta}}(\cdot)$ should fit $p_\text{data}(\cdot)$ with a maximized likelihood $\mathcal{L}(\mathcal{E}_{\boldsymbol{\theta}}, \mathbf{X})$.
    Here, 
    our likelihood function adopts the common expected log-likelihood over the data distribution in the EBM framework as
    \begin{equation}
        \label{eq:ll}
        \mathcal{L}(\mathcal{E}_{\boldsymbol{\theta}}, \mathbf{X}) = \mathbb{E}_{\mathbf{x}\sim p_\text{data}} [\log Q_{\boldsymbol{\theta}}(\mathbf{x})]=\mathbb{E}[\log \frac{1}{Z_{\boldsymbol{\theta}}}\exp(-\mathcal{E}_{\boldsymbol{\theta}}(\mathbf{x}))],
    \end{equation}
    where $\boldsymbol{\theta}$ denotes the network parameters, 
    and $Z_{\boldsymbol{\theta}}$ is the partition function dependent on $\boldsymbol{\theta}$.
    Then, 
    we can train $\mathcal{E}_{\boldsymbol{\theta}}$ to maximize (\ref{eq:ll}).
    A common training methodology is to define a loss function as the negative log-likelihood (NLL):
    \begin{equation}
        \label{eq:nll}
        L_{\boldsymbol{\theta}} = -\mathcal{L}(\mathcal{E}_{\boldsymbol{\theta}}, \mathbf{X})=\frac{1}{N}\sum\nolimits_{i=1}^{N}(\mathcal{E}_{\boldsymbol{\theta}}(\mathbf{x}_i) + \log(Z_{\boldsymbol{\theta}})),
    \end{equation}
    and adopt the Markov Chain Monte Carlo (MCMC) method with Langevin sampling~\cite{parisi1981correlation,grenander1994representations}.
    While computing the partition function $Z_{\boldsymbol{\theta}}$ itself is intractable,
    its gradient $\nabla_{\boldsymbol{\theta}} \log Z_{\boldsymbol{\theta}}$ can be efficiently estimated via MCMC sampling,
    which is sufficient for gradient-based optimization (see SI for detailed methodology and parameter setups).
    Crucially, 
    this approach discovers probability distributions directly from raw data points via likelihood maximization, 
    rather than fitting predefined probability density functions (PDFs) to pre-computed statistics.

    \subsubsection{Extracting intermediate representations by inference}
    The post-trained Deflexformer $\mathcal{E}_{\boldsymbol{\theta}}$ approximates an energy function $E(\cdot)$ while the post-trained block $\mathcal{E}^i$ corresponds to the base functions $E^i(\cdot)$ within $E(\cdot)$.
    As $P_i$ and $G_i$ denote the element-level and global-level representations output by the $\mathcal{E}^i$ ($P_0$ and $G_0$ correspond to the representations of raw data $\mathbf{X}$), there should be $P_i||G_i=\mathcal{E}^i(P_{i-1}||G_{i-1})$ where $P_i||G_i$ means the concatenation of representations.
    More generally, there is $P_i||G_i= \mathcal{E}^{i-l} \circ \cdots \circ \mathcal{E}^i(P_{i-l-1}||G_{i-l-1})$.
    Therefore, to identify any relationships $E^{i-l} \circ \cdots \circ E^i$, we can perform SR on a large number of input-output representation pairs $(P_{i-l-1}||G_{i-l-1}, P_i||G_i)$ by inferring over the post-trained Deflexformer $\mathcal{E}_{\boldsymbol{\theta}}$.
    Here, we can feed each sample in the observation data into Deflexformer to conduct forward computation over a block array $\mathcal{E}^{i-l} \circ \cdots \circ \mathcal{E}^i$ of multiple sequential blocks, producing the corresponding intermediate representations.
    After multiple inferences, we can collect a dataset to perform SR, thus identifying any relationship represented by an array with an arbitrary number of sequential blocks.
    However, the number of collected samples may be constrained by that of the observation dataset, thereby limiting SR accuracy.
    Fortunately, as an energy model, Deflexformer inherently supports generating more data samples with a similar distribution to the original observation data.
    Specifically, it can work by using Langevin sampling from random noise (see SI for sampling details).

    \subsubsection{Hierarchical SR from intermediate representations}
    SR for a smaller block array with fewer Deflexformer blocks can reduce the computation complexity, but may converge to fragmented expressions with no meaning or oversimplified with larger decomposition error.
    Then their composition would lead to significantly biased energy functions.
    Nevertheless, these incomplete/inaccurate expressions can serve as base expressions to accelerate the SR for a larger block array, which has smaller decomposition errors.
    To this end, we propose a hierarchical SR strategy for recovering the whole energy function.
    Here, denote $\hat{E}^{i-l}\circ\ldots \circ \hat{E}^k\leftarrow SR(P_{i-l-1}||G_{i-l-1}, P_k||G_k)$ as SR for the $l$ consecutive blocks indexed by $k-l, k-l+1, \ldots, k$.
    Let $k=1,2,\ldots, K$ denote the block index.
    We perform the SRs by building a binary tree $\mathcal{T}$ with the depth of $\lceil \log K \rceil$.
    For each node $v$ in $\mathcal{T}$, we denote $\mathcal{T}(v)$ as the sub-tree with node $v$ as its root.
    In $\mathcal{T}$, each node $v$ represents an SR process over the consecutive blocks indexed by the leaf nodes of the subtree $\mathcal{T}(v)$.
    The data samples (input-output representation pairs) for $SR_v$ are constructed by extracting the pair of the input representation of $\mathcal{T}(v)$'s leftmost leaf node and the corresponding output representation of $\mathcal{T}(v)$'s rightmost leaf node.
    In particular, for the leaf layer, the base expression set for SR is user-specified.
    For other layers in the tree, the base expression set for any node $v$ is constructed by taking the union of the SR-derived expression set of its two children.
    Up to the root, we can gradually refine the SR results and finally obtain an expression set $\{\hat{E}_1,\ldots,\hat{E}_m\}$ for the entire Deflexformer that approximates the complete energy function $E(\cdot)$.
    Although this incurs $O(K\log K)$ SRs, it can greatly reduce the searching space of SR for complex relationships (see SI for detailed algorithm). 
    
    \section*{Data availability}
    
    The data generated or processed in this study are available in the open-source project repository at https://github.com/yhqjohn/deflex. For the simulation-based projects, the repository provides the generated datasets for rare gas particle motion, pollen-in-water particle motion, cross-scale water particles, and two-dimensional cylinder-wake fluid dynamics. The original third-party datasets used in this study are publicly available from their source repositories or publications: the Zurich Carnival human-mobility dataset at https://doi.org/10.4108/icst.urb-iot.2014.257190; the eBird Status and Trends Data Version 2022 at https://doi.org/10.2173/ebirdst.2022; the JHTDB dataset at https://turbulence.pha.jhu.edu/; and the Feynman Symbolic Regression Database at https://space.mit.edu/home/tegmark/aifeynman.html, with associated publication DOI https://doi.org/10.1126/sciadv.aay2631. No additional access restrictions apply to the data made available through the project repository.

    
    \section*{Code availability}
    
    The code used in the study is available in the open-source development repository at https://github.com/yhqjohn/deflex. 
    
    
    \bigskip
    
    

\begin{thebibliography}{54}
    \ifx \bisbn   \undefined \def \bisbn  #1{ISBN #1}\fi
    \ifx \binits  \undefined \def \binits#1{#1}\fi
    \ifx \bauthor  \undefined \def \bauthor#1{#1}\fi
    \ifx \batitle  \undefined \def \batitle#1{#1}\fi
    \ifx \bjtitle  \undefined \def \bjtitle#1{#1}\fi
    \ifx \bvolume  \undefined \def \bvolume#1{\textbf{#1}}\fi
    \ifx \byear  \undefined \def \byear#1{#1}\fi
    \ifx \bissue  \undefined \def \bissue#1{#1}\fi
    \ifx \bfpage  \undefined \def \bfpage#1{#1}\fi
    \ifx \blpage  \undefined \def \blpage #1{#1}\fi
    \ifx \burl  \undefined \def \burl#1{\textsf{#1}}\fi
    \ifx \doiurl  \undefined \def \doiurl#1{\url{https://doi.org/#1}}\fi
    \ifx \betal  \undefined \def \betal{\textit{et al.}}\fi
    \ifx \binstitute  \undefined \def \binstitute#1{#1}\fi
    \ifx \binstitutionaled  \undefined \def \binstitutionaled#1{#1}\fi
    \ifx \bctitle  \undefined \def \bctitle#1{#1}\fi
    \ifx \beditor  \undefined \def \beditor#1{#1}\fi
    \ifx \bpublisher  \undefined \def \bpublisher#1{#1}\fi
    \ifx \bbtitle  \undefined \def \bbtitle#1{#1}\fi
    \ifx \bedition  \undefined \def \bedition#1{#1}\fi
    \ifx \bseriesno  \undefined \def \bseriesno#1{#1}\fi
    \ifx \blocation  \undefined \def \blocation#1{#1}\fi
    \ifx \bsertitle  \undefined \def \bsertitle#1{#1}\fi
    \ifx \bsnm \undefined \def \bsnm#1{#1}\fi
    \ifx \bsuffix \undefined \def \bsuffix#1{#1}\fi
    \ifx \bparticle \undefined \def \bparticle#1{#1}\fi
    \ifx \barticle \undefined \def \barticle#1{#1}\fi
    \bibcommenthead
    \ifx \bconfdate \undefined \def \bconfdate #1{#1}\fi
    \ifx \botherref \undefined \def \botherref #1{#1}\fi
    \ifx \url \undefined \def \url#1{\textsf{#1}}\fi
    \ifx \bchapter \undefined \def \bchapter#1{#1}\fi
    \ifx \bbook \undefined \def \bbook#1{#1}\fi
    \ifx \bcomment \undefined \def \bcomment#1{#1}\fi
    \ifx \oauthor \undefined \def \oauthor#1{#1}\fi
    \ifx \citeauthoryear \undefined \def \citeauthoryear#1{#1}\fi
    \ifx \endbibitem  \undefined \def \endbibitem {}\fi
    \ifx \bconflocation  \undefined \def \bconflocation#1{#1}\fi
    \ifx \arxivurl  \undefined \def \arxivurl#1{\textsf{#1}}\fi
    \csname PreBibitemsHook\endcsname
    
    \bibitem[\protect\citeauthoryear{Wigner}{1990}]{wigner1990unreasonable}
    \begin{bchapter}
    \bauthor{\bsnm{Wigner}, \binits{E.P.}}:
    \bctitle{The unreasonable effectiveness of mathematics in the natural sciences}.
    In: \bbtitle{Mathematics and Science},
    pp. \bfpage{291}--\blpage{306}.
    \bpublisher{World Scientific},
    \blocation{Singapore}
    (\byear{1990})
    \end{bchapter}
    \endbibitem
    
    \bibitem[\protect\citeauthoryear{Schmidt and Lipson}{2009}]{doi:10.1126/science.1165893}
    \begin{barticle}
    \bauthor{\bsnm{Schmidt}, \binits{M.}},
    \bauthor{\bsnm{Lipson}, \binits{H.}}:
    \batitle{Distilling free-form natural laws from experimental data}.
    \bjtitle{Science}
    \bvolume{324}(\bissue{5923}),
    \bfpage{81}--\blpage{85}
    (\byear{2009})
    \doiurl{10.1126/science.1165893}
    \end{barticle}
    \endbibitem
    
    \bibitem[\protect\citeauthoryear{Irrgang et~al.}{2021}]{irrgang2021towards}
    \begin{barticle}
    \bauthor{\bsnm{Irrgang}, \binits{C.}},
    \bauthor{\bsnm{Boers}, \binits{N.}},
    \bauthor{\bsnm{Sonnewald}, \binits{M.}},
    \bauthor{\bsnm{Barnes}, \binits{E.A.}},
    \bauthor{\bsnm{Kadow}, \binits{C.}},
    \bauthor{\bsnm{Staneva}, \binits{J.}},
    \bauthor{\bsnm{Saynisch-Wagner}, \binits{J.}}:
    \batitle{Towards neural earth system modelling by integrating artificial intelligence in earth system science}.
    \bjtitle{Nature Machine Intelligence}
    \bvolume{3}(\bissue{8}),
    \bfpage{667}--\blpage{674}
    (\byear{2021})
    \end{barticle}
    \endbibitem
    
    \bibitem[\protect\citeauthoryear{Qi and Majda}{2020}]{doi:10.1073/pnas.1917285117}
    \begin{barticle}
    \bauthor{\bsnm{Qi}, \binits{D.}},
    \bauthor{\bsnm{Majda}, \binits{A.J.}}:
    \batitle{Using machine learning to predict extreme events in complex systems}.
    \bjtitle{Proceedings of the National Academy of Sciences}
    \bvolume{117}(\bissue{1}),
    \bfpage{52}--\blpage{59}
    (\byear{2020})
    \doiurl{10.1073/pnas.1917285117}
    \end{barticle}
    \endbibitem
    
    \bibitem[\protect\citeauthoryear{Varadi et~al.}{2022}]{varadi2022alphafold}
    \begin{barticle}
    \bauthor{\bsnm{Varadi}, \binits{M.}},
    \bauthor{\bsnm{Anyango}, \binits{S.}},
    \bauthor{\bsnm{Deshpande}, \binits{M.}},
    \bauthor{\bsnm{Nair}, \binits{S.}},
    \bauthor{\bsnm{Natassia}, \binits{C.}},
    \bauthor{\bsnm{Yordanova}, \binits{G.}},
    \bauthor{\bsnm{Yuan}, \binits{D.}},
    \bauthor{\bsnm{Stroe}, \binits{O.}},
    \bauthor{\bsnm{Wood}, \binits{G.}},
    \bauthor{\bsnm{Laydon}, \binits{A.}}, \betal:
    \batitle{Alphafold protein structure database: massively expanding the structural coverage of protein-sequence space with high-accuracy models}.
    \bjtitle{Nucleic acids research}
    \bvolume{50}(\bissue{D1}),
    \bfpage{439}--\blpage{444}
    (\byear{2022})
    \end{barticle}
    \endbibitem
    
    \bibitem[\protect\citeauthoryear{Udrescu et~al.}{2020}]{udrescu2020ai}
    \begin{barticle}
    \bauthor{\bsnm{Udrescu}, \binits{S.-M.}},
    \bauthor{\bsnm{Tan}, \binits{A.}},
    \bauthor{\bsnm{Feng}, \binits{J.}},
    \bauthor{\bsnm{Neto}, \binits{O.}},
    \bauthor{\bsnm{Wu}, \binits{T.}},
    \bauthor{\bsnm{Tegmark}, \binits{M.}}:
    \batitle{Ai feynman 2.0: Pareto-optimal symbolic regression exploiting graph modularity}.
    \bjtitle{Advances in Neural Information Processing Systems}
    \bvolume{33},
    \bfpage{4860}--\blpage{4871}
    (\byear{2020})
    \end{barticle}
    \endbibitem
    
    \bibitem[\protect\citeauthoryear{Udrescu and Tegmark}{2020}]{doi:10.1126/sciadv.aay2631}
    \begin{barticle}
    \bauthor{\bsnm{Udrescu}, \binits{S.-M.}},
    \bauthor{\bsnm{Tegmark}, \binits{M.}}:
    \batitle{Ai feynman: A physics-inspired method for symbolic regression}.
    \bjtitle{Science Advances}
    \bvolume{6}(\bissue{16}),
    \bfpage{2631}
    (\byear{2020})
    \doiurl{10.1126/sciadv.aay2631}
    \end{barticle}
    \endbibitem
    
    \bibitem[\protect\citeauthoryear{Bellomo and Dogbe}{2011}]{bellomo2011modeling}
    \begin{barticle}
    \bauthor{\bsnm{Bellomo}, \binits{N.}},
    \bauthor{\bsnm{Dogbe}, \binits{C.}}:
    \batitle{On the modeling of traffic and crowds: A survey of models, speculations, and perspectives}.
    \bjtitle{SIAM review}
    \bvolume{53}(\bissue{3}),
    \bfpage{409}--\blpage{463}
    (\byear{2011})
    \end{barticle}
    \endbibitem
    
    \bibitem[\protect\citeauthoryear{Bar-Yam}{2002}]{bar2002general}
    \begin{botherref}
    \oauthor{\bsnm{Bar-Yam}, \binits{Y.}}:
    General features of complex systems.
    Encyclopedia of Life Support Systems (EOLSS), UNESCO, EOLSS Publishers, Oxford, UK
    \textbf{1}
    (2002)
    \end{botherref}
    \endbibitem
    
    \bibitem[\protect\citeauthoryear{Council et~al.}{2012}]{national2012national}
    \begin{bbook}
    \bauthor{\bsnm{Council}, \binits{N.R.}}, \betal:
    \bbtitle{A National Strategy for Advancing Climate Modeling}.
    \bpublisher{MIT Press},
    \blocation{Cambridge, MA}
    (\byear{2012})
    \end{bbook}
    \endbibitem
    
    \bibitem[\protect\citeauthoryear{Fortunato et~al.}{2018}]{fortunato2018science}
    \begin{barticle}
    \bauthor{\bsnm{Fortunato}, \binits{S.}},
    \bauthor{\bsnm{Bergstrom}, \binits{C.T.}},
    \bauthor{\bsnm{B{\"o}rner}, \binits{K.}},
    \bauthor{\bsnm{Evans}, \binits{J.A.}},
    \bauthor{\bsnm{Helbing}, \binits{D.}},
    \bauthor{\bsnm{Milojevi{\'c}}, \binits{S.}},
    \bauthor{\bsnm{Petersen}, \binits{A.M.}},
    \bauthor{\bsnm{Radicchi}, \binits{F.}},
    \bauthor{\bsnm{Sinatra}, \binits{R.}},
    \bauthor{\bsnm{Uzzi}, \binits{B.}}, \betal:
    \batitle{Science of science}.
    \bjtitle{Science}
    \bvolume{359}(\bissue{6379}),
    \bfpage{0185}
    (\byear{2018})
    \end{barticle}
    \endbibitem
    
    \bibitem[\protect\citeauthoryear{Pomeau}{2016}]{pomeau2016long}
    \begin{barticle}
    \bauthor{\bsnm{Pomeau}, \binits{Y.}}:
    \batitle{The long and winding road}.
    \bjtitle{Nature Physics}
    \bvolume{12}(\bissue{3}),
    \bfpage{198}--\blpage{199}
    (\byear{2016})
    \end{barticle}
    \endbibitem
    
    \bibitem[\protect\citeauthoryear{Cranmer et~al.}{2020}]{10.5555/3495724.3497186}
    \begin{bchapter}
    \bauthor{\bsnm{Cranmer}, \binits{M.}},
    \bauthor{\bsnm{Sanchez-Gonzalez}, \binits{A.}},
    \bauthor{\bsnm{Battaglia}, \binits{P.}},
    \bauthor{\bsnm{Xu}, \binits{R.}},
    \bauthor{\bsnm{Cranmer}, \binits{K.}},
    \bauthor{\bsnm{Spergel}, \binits{D.}},
    \bauthor{\bsnm{Ho}, \binits{S.}}:
    \bctitle{Discovering symbolic models from deep learning with inductive biases}.
    In: \bbtitle{Proceedings of the 34th International Conference on Neural Information Processing Systems}.
    \bsertitle{NIPS'20}.
    \bpublisher{Curran Associates Inc.},
    \blocation{Red Hook, NY, USA}
    (\byear{2020})
    \end{bchapter}
    \endbibitem
    
    \bibitem[\protect\citeauthoryear{Rudy et~al.}{2017}]{doi:10.1126/sciadv.1602614}
    \begin{barticle}
    \bauthor{\bsnm{Rudy}, \binits{S.H.}},
    \bauthor{\bsnm{Brunton}, \binits{S.L.}},
    \bauthor{\bsnm{Proctor}, \binits{J.L.}},
    \bauthor{\bsnm{Kutz}, \binits{J.N.}}:
    \batitle{Data-driven discovery of partial differential equations}.
    \bjtitle{Science Advances}
    \bvolume{3}(\bissue{4}),
    \bfpage{1602614}
    (\byear{2017})
    \doiurl{10.1126/sciadv.1602614}
    \end{barticle}
    \endbibitem
    
    \bibitem[\protect\citeauthoryear{Gao et~al.}{2024}]{gao2024learning}
    \begin{barticle}
    \bauthor{\bsnm{Gao}, \binits{T.-T.}},
    \bauthor{\bsnm{Barzel}, \binits{B.}},
    \bauthor{\bsnm{Yan}, \binits{G.}}:
    \batitle{Learning interpretable dynamics of stochastic complex systems from experimental data}.
    \bjtitle{Nature Communications}
    \bvolume{15}(\bissue{1}),
    \bfpage{6029}
    (\byear{2024})
    \end{barticle}
    \endbibitem
    
    \bibitem[\protect\citeauthoryear{Van~Doren and Horton}{2018}]{van2018continental}
    \begin{barticle}
    \bauthor{\bsnm{Van~Doren}, \binits{B.M.}},
    \bauthor{\bsnm{Horton}, \binits{K.G.}}:
    \batitle{A continental system for forecasting bird migration}.
    \bjtitle{Science}
    \bvolume{361}(\bissue{6407}),
    \bfpage{1115}--\blpage{1118}
    (\byear{2018})
    \end{barticle}
    \endbibitem
    
    \bibitem[\protect\citeauthoryear{Hasselmann}{1976}]{hasselmann1976stochastic}
    \begin{barticle}
    \bauthor{\bsnm{Hasselmann}, \binits{K.}}:
    \batitle{Stochastic climate models part i. theory}.
    \bjtitle{tellus}
    \bvolume{28}(\bissue{6}),
    \bfpage{473}--\blpage{485}
    (\byear{1976})
    \end{barticle}
    \endbibitem
    
    \bibitem[\protect\citeauthoryear{Vicsek}{2012}]{vicsek2012collective}
    \begin{botherref}
    \oauthor{\bsnm{Vicsek}, \binits{T.}}:
    Collective Motion.
    Elsevier
    (2012)
    \end{botherref}
    \endbibitem
    
    \bibitem[\protect\citeauthoryear{Kaneko and Tsuda}{2001}]{kaneko2001complex}
    \begin{bbook}
    \bauthor{\bsnm{Kaneko}, \binits{K.}},
    \bauthor{\bsnm{Tsuda}, \binits{I.}}:
    \bbtitle{Complex Systems: Chaos and Beyond: Chaos and Beyond: A Constructive Approach with Applications in Life Sciences}.
    \bpublisher{Springer},
    \blocation{Berlin/Heidelberg, Germany}
    (\byear{2001})
    \end{bbook}
    \endbibitem
    
    \bibitem[\protect\citeauthoryear{Weinan}{2011}]{weinan2011principles}
    \begin{bbook}
    \bauthor{\bsnm{Weinan}, \binits{E.}}:
    \bbtitle{Principles of Multiscale Modeling}.
    \bpublisher{Cambridge University Press},
    \blocation{Cambridge, UK}
    (\byear{2011})
    \end{bbook}
    \endbibitem
    
    \bibitem[\protect\citeauthoryear{Pun et~al.}{2019}]{pun2019physically}
    \begin{barticle}
    \bauthor{\bsnm{Pun}, \binits{G.P.}},
    \bauthor{\bsnm{Batra}, \binits{R.}},
    \bauthor{\bsnm{Ramprasad}, \binits{R.}},
    \bauthor{\bsnm{Mishin}, \binits{Y.}}:
    \batitle{Physically informed artificial neural networks for atomistic modeling of materials}.
    \bjtitle{Nature communications}
    \bvolume{10}(\bissue{1}),
    \bfpage{2339}
    (\byear{2019})
    \end{barticle}
    \endbibitem
    
    \bibitem[\protect\citeauthoryear{Bi et~al.}{2023}]{bi2023accurate}
    \begin{barticle}
    \bauthor{\bsnm{Bi}, \binits{K.}},
    \bauthor{\bsnm{Xie}, \binits{L.}},
    \bauthor{\bsnm{Zhang}, \binits{H.}},
    \bauthor{\bsnm{Chen}, \binits{X.}},
    \bauthor{\bsnm{Gu}, \binits{X.}},
    \bauthor{\bsnm{Tian}, \binits{Q.}}:
    \batitle{Accurate medium-range global weather forecasting with 3d neural networks}.
    \bjtitle{Nature}
    \bvolume{619}(\bissue{7970}),
    \bfpage{533}--\blpage{538}
    (\byear{2023})
    \end{barticle}
    \endbibitem
    
    \bibitem[\protect\citeauthoryear{Romera-Paredes et~al.}{2024}]{romera2024mathematical}
    \begin{barticle}
    \bauthor{\bsnm{Romera-Paredes}, \binits{B.}},
    \bauthor{\bsnm{Barekatain}, \binits{M.}},
    \bauthor{\bsnm{Novikov}, \binits{A.}},
    \bauthor{\bsnm{Balog}, \binits{M.}},
    \bauthor{\bsnm{Kumar}, \binits{M.P.}},
    \bauthor{\bsnm{Dupont}, \binits{E.}},
    \bauthor{\bsnm{Ruiz}, \binits{F.J.}},
    \bauthor{\bsnm{Ellenberg}, \binits{J.S.}},
    \bauthor{\bsnm{Wang}, \binits{P.}},
    \bauthor{\bsnm{Fawzi}, \binits{O.}}, \betal:
    \batitle{Mathematical discoveries from program search with large language models}.
    \bjtitle{Nature}
    \bvolume{625}(\bissue{7995}),
    \bfpage{468}--\blpage{475}
    (\byear{2024})
    \end{barticle}
    \endbibitem
    
    \bibitem[\protect\citeauthoryear{Brunton et~al.}{2016}]{doi:10.1073/pnas.1517384113}
    \begin{barticle}
    \bauthor{\bsnm{Brunton}, \binits{S.L.}},
    \bauthor{\bsnm{Proctor}, \binits{J.L.}},
    \bauthor{\bsnm{Kutz}, \binits{J.N.}}:
    \batitle{Discovering governing equations from data by sparse identification of nonlinear dynamical systems}.
    \bjtitle{Proceedings of the National Academy of Sciences}
    \bvolume{113}(\bissue{15}),
    \bfpage{3932}--\blpage{3937}
    (\byear{2016})
    \doiurl{10.1073/pnas.1517384113}
    \end{barticle}
    \endbibitem
    
    \bibitem[\protect\citeauthoryear{Koza}{1994}]{koza1994genetic}
    \begin{barticle}
    \bauthor{\bsnm{Koza}, \binits{J.R.}}:
    \batitle{Genetic programming as a means for programming computers by natural selection}.
    \bjtitle{Statistics and computing}
    \bvolume{4},
    \bfpage{87}--\blpage{112}
    (\byear{1994})
    \end{barticle}
    \endbibitem
    
    \bibitem[\protect\citeauthoryear{Keren et~al.}{2023}]{keren2023computational}
    \begin{barticle}
    \bauthor{\bsnm{Keren}, \binits{L.S.}},
    \bauthor{\bsnm{Liberzon}, \binits{A.}},
    \bauthor{\bsnm{Lazebnik}, \binits{T.}}:
    \batitle{A computational framework for physics-informed symbolic regression with straightforward integration of domain knowledge}.
    \bjtitle{Scientific Reports}
    \bvolume{13}(\bissue{1}),
    \bfpage{1249}
    (\byear{2023})
    \end{barticle}
    \endbibitem
    
    \bibitem[\protect\citeauthoryear{Sun et~al.}{2023}]{sun2023symbolic}
    \begin{bchapter}
    \bauthor{\bsnm{Sun}, \binits{F.}},
    \bauthor{\bsnm{Liu}, \binits{Y.}},
    \bauthor{\bsnm{Wang}, \binits{J.-X.}},
    \bauthor{\bsnm{Sun}, \binits{H.}}:
    \bctitle{Symbolic physics learner: Discovering governing equations via monte carlo tree search}.
    In: \bbtitle{The Eleventh International Conference on Learning Representations}
    (\byear{2023})
    \end{bchapter}
    \endbibitem
    
    \bibitem[\protect\citeauthoryear{Jiang and Xue}{2024}]{10.1609/aaai.v38i11.29187}
    \begin{bchapter}
    \bauthor{\bsnm{Jiang}, \binits{N.}},
    \bauthor{\bsnm{Xue}, \binits{Y.}}:
    \bctitle{Racing control variable genetic programming for symbolic regression}.
    In: \bbtitle{Proceedings of the 38th AAAI Conference on Artificial Intelligence (AAAI 2024)}.
    \bpublisher{AAAI Press},
    \blocation{Vancouver, Canada}
    (\byear{2024}).
    \doiurl{10.1609/aaai.v38i11.29187}
    \end{bchapter}
    \endbibitem
    
    \bibitem[\protect\citeauthoryear{Mundhenk et~al.}{2021}]{NEURIPS2021_d073bb8d}
    \begin{bchapter}
    \bauthor{\bsnm{Mundhenk}, \binits{T.}},
    \bauthor{\bsnm{Landajuela}, \binits{M.}},
    \bauthor{\bsnm{Glatt}, \binits{R.}},
    \bauthor{\bsnm{Santiago}, \binits{C.P.}},
    \bauthor{\bsnm{faissol}, \binits{D.}},
    \bauthor{\bsnm{Petersen}, \binits{B.K.}}:
    \bctitle{Symbolic regression via deep reinforcement learning enhanced genetic programming seeding}.
    In: \beditor{\bsnm{Ranzato}, \binits{M.}},
    \beditor{\bsnm{Beygelzimer}, \binits{A.}},
    \beditor{\bsnm{Dauphin}, \binits{Y.}},
    \beditor{\bsnm{Liang}, \binits{P.S.}},
    \beditor{\bsnm{Vaughan}, \binits{J.W.}} (eds.)
    \bbtitle{Advances in Neural Information Processing Systems},
    vol. \bseriesno{34},
    pp. \bfpage{24912}--\blpage{24923}.
    \bpublisher{Curran Associates, Inc.},
    \blocation{Second Virtual Conference}
    (\byear{2021})
    \end{bchapter}
    \endbibitem
    
    \bibitem[\protect\citeauthoryear{Siegenfeld and Bar-Yam}{2020}]{siegenfeld2020introduction}
    \begin{barticle}
    \bauthor{\bsnm{Siegenfeld}, \binits{A.F.}},
    \bauthor{\bsnm{Bar-Yam}, \binits{Y.}}:
    \batitle{An introduction to complex systems science and its applications}.
    \bjtitle{Complexity}
    \bvolume{2020},
    \bfpage{1}--\blpage{16}
    (\byear{2020})
    \end{barticle}
    \endbibitem
    
    \bibitem[\protect\citeauthoryear{Burlacu et~al.}{2020}]{10.1145/3377929.3398099}
    \begin{bchapter}
    \bauthor{\bsnm{Burlacu}, \binits{B.}},
    \bauthor{\bsnm{Kronberger}, \binits{G.}},
    \bauthor{\bsnm{Kommenda}, \binits{M.}}:
    \bctitle{Operon c++: An efficient genetic programming framework for symbolic regression}.
    In: \bbtitle{Proceedings of the 2020 Genetic and Evolutionary Computation Conference Companion}.
    \bsertitle{GECCO '20},
    pp. \bfpage{1562}--\blpage{1570}.
    \bpublisher{Association for Computing Machinery},
    \blocation{New York, NY, USA}
    (\byear{2020}).
    \doiurl{10.1145/3377929.3398099}
    \end{bchapter}
    \endbibitem
    
    \bibitem[\protect\citeauthoryear{Stephens et~al.}{2025}]{Stephens2025trevorstephens}
    \begin{botherref}
    \oauthor{\bsnm{Stephens}, \binits{T.}},
    \oauthor{\bsnm{Kemenade}, \binits{H.}},
    \oauthor{\bsnm{Rai}, \binits{A.}},
    \oauthor{\bsnm{Price}, \binits{B.}},
    \oauthor{\bsnm{Watson}, \binits{C.}},
    \oauthor{\bsnm{Bell}, \binits{I.}},
    \oauthor{\bsnm{McDermott}, \binits{J.}},
    \oauthor{\bsnm{Vos}, \binits{N.}},
    \oauthor{\bsnm{Niculae}, \binits{S.}},
    \oauthor{\bsnm{Di~Caprio}, \binits{U.}},
    \oauthor{\bsnm{{sun ao}}}:
    trevorstephens/gplearn.
    https://github.com/trevorstephens/gplearn
    (2025).
    \url{https://github.com/trevorstephens/gplearn}
    \end{botherref}
    \endbibitem
    
    \bibitem[\protect\citeauthoryear{Cranmer}{2023}]{cranmerInterpretableMachineLearning2023}
    \begin{botherref}
    \oauthor{\bsnm{Cranmer}, \binits{M.}}:
    Interpretable {Machine} {Learning} for {Science} with {PySR} and {SymbolicRegression}.jl.
    arXiv
    (2023).
    \doiurl{10.48550/arXiv.2305.01582}
    \end{botherref}
    \endbibitem
    
    \bibitem[\protect\citeauthoryear{Fortin et~al.}{2012}]{DEAP_JMLR2012}
    \begin{barticle}
    \bauthor{\bsnm{Fortin}, \binits{F.-A.}},
    \bauthor{\bsnm{{De Rainville}}, \binits{F.-M.}},
    \bauthor{\bsnm{Gardner}, \binits{M.-A.}},
    \bauthor{\bsnm{Parizeau}, \binits{M.}},
    \bauthor{\bsnm{Gagn\'e}, \binits{C.}}:
    \batitle{{DEAP}: Evolutionary algorithms made easy}.
    \bjtitle{Journal of Machine Learning Research}
    \bvolume{13},
    \bfpage{2171}--\blpage{2175}
    (\byear{2012})
    \end{barticle}
    \endbibitem
    
    \bibitem[\protect\citeauthoryear{Villegas et~al.}{2023}]{villegas2023laplacian}
    \begin{barticle}
    \bauthor{\bsnm{Villegas}, \binits{P.}},
    \bauthor{\bsnm{Gili}, \binits{T.}},
    \bauthor{\bsnm{Caldarelli}, \binits{G.}},
    \bauthor{\bsnm{Gabrielli}, \binits{A.}}:
    \batitle{Laplacian renormalization group for heterogeneous networks}.
    \bjtitle{Nature Physics}
    \bvolume{19}(\bissue{3}),
    \bfpage{445}--\blpage{450}
    (\byear{2023})
    \end{barticle}
    \endbibitem
    
    \bibitem[\protect\citeauthoryear{Villani et~al.}{2009}]{villani2009optimal}
    \begin{bbook}
    \bauthor{\bsnm{Villani}, \binits{C.}}, \betal:
    \bbtitle{Optimal Transport: Old and New}.
    \bsertitle{Grundlehren der mathematischen Wissenschaften},
    vol. \bseriesno{338}.
    \bpublisher{Springer},
    \blocation{Berlin Heidelberg}
    (\byear{2009})
    \end{bbook}
    \endbibitem
    
    \bibitem[\protect\citeauthoryear{Langevin et~al.}{1908}]{langevin1908theorie}
    \begin{barticle}
    \bauthor{\bsnm{Langevin}, \binits{P.}}, \betal:
    \batitle{Sur la th{\'e}orie du mouvement brownien}.
    \bjtitle{CR Acad. Sci. Paris}
    \bvolume{146}(\bissue{530-533}),
    \bfpage{530}
    (\byear{1908})
    \end{barticle}
    \endbibitem
    
    \bibitem[\protect\citeauthoryear{Blanke et~al.}{2014}]{10.4108/icst.urb-iot.2014.257190}
    \begin{bchapter}
    \bauthor{\bsnm{Blanke}, \binits{U.}},
    \bauthor{\bsnm{Guldener}, \binits{R.}},
    \bauthor{\bsnm{Feese}, \binits{S.}},
    \bauthor{\bsnm{Tr\"{o}ster}, \binits{G.}}:
    \bctitle{Crowdsourced pedestrian map construction for short-term city-scale events}.
    In: \bbtitle{Proceedings of the First International Conference on IoT in Urban Space}.
    \bsertitle{URB-IOT '14},
    pp. \bfpage{25}--\blpage{31}.
    \bpublisher{ICST (Institute for Computer Sciences, Social-Informatics and Telecommunications Engineering)},
    \blocation{Brussels, BEL}
    (\byear{2014}).
    \doiurl{10.4108/icst.urb-iot.2014.257190}
    \end{bchapter}
    \endbibitem
    
    \bibitem[\protect\citeauthoryear{Fink et~al.}{2023}]{eBirdStatusTrends2022}
    \begin{botherref}
    \oauthor{\bsnm{Fink}, \binits{D.}},
    \oauthor{\bsnm{Auer}, \binits{T.}},
    \oauthor{\bsnm{Johnston}, \binits{A.}},
    \oauthor{\bsnm{Strimas-Mackey}, \binits{M.}},
    \oauthor{\bsnm{Ligocki}, \binits{S.}},
    \oauthor{\bsnm{Robinson}, \binits{O.}},
    \oauthor{\bsnm{Hochachka}, \binits{W.}},
    \oauthor{\bsnm{Jaromczyk}, \binits{L.}},
    \oauthor{\bsnm{Crowlye}, \binits{C.}},
    \oauthor{\bsnm{Dunham}, \binits{K.}},
    \oauthor{\bsnm{Stillman}, \binits{A.}},
    \oauthor{\bsnm{Davies}, \binits{I.}},
    \oauthor{\bsnm{Rodewald}, \binits{A.}},
    \oauthor{\bsnm{Ruiz-Gutierrez}, \binits{V.}},
    \oauthor{\bsnm{Wood}, \binits{C.}}:
    eBird Status and Trends, Data Version: 2022; Released: 2023.
    Cornell Lab of Ornithology
    (2023).
    \doiurl{10.2173/ebirdst.2022}
    \end{botherref}
    \endbibitem
    
    \bibitem[\protect\citeauthoryear{Gonzalez et~al.}{2008}]{gonzalez2008understanding}
    \begin{barticle}
    \bauthor{\bsnm{Gonzalez}, \binits{M.C.}},
    \bauthor{\bsnm{Hidalgo}, \binits{C.A.}},
    \bauthor{\bsnm{Barabasi}, \binits{A.-L.}}:
    \batitle{Understanding individual human mobility patterns}.
    \bjtitle{nature}
    \bvolume{453}(\bissue{7196}),
    \bfpage{779}--\blpage{782}
    (\byear{2008})
    \end{barticle}
    \endbibitem
    
    \bibitem[\protect\citeauthoryear{Schl{\"a}pfer et~al.}{2021}]{schlapfer2021universal}
    \begin{barticle}
    \bauthor{\bsnm{Schl{\"a}pfer}, \binits{M.}},
    \bauthor{\bsnm{Dong}, \binits{L.}},
    \bauthor{\bsnm{O’Keeffe}, \binits{K.}},
    \bauthor{\bsnm{Santi}, \binits{P.}},
    \bauthor{\bsnm{Szell}, \binits{M.}},
    \bauthor{\bsnm{Salat}, \binits{H.}},
    \bauthor{\bsnm{Anklesaria}, \binits{S.}},
    \bauthor{\bsnm{Vazifeh}, \binits{M.}},
    \bauthor{\bsnm{Ratti}, \binits{C.}},
    \bauthor{\bsnm{West}, \binits{G.B.}}:
    \batitle{The universal visitation law of human mobility}.
    \bjtitle{Nature}
    \bvolume{593}(\bissue{7860}),
    \bfpage{522}--\blpage{527}
    (\byear{2021})
    \end{barticle}
    \endbibitem
    
    \bibitem[\protect\citeauthoryear{Bongiorno et~al.}{2021}]{bongiorno2021vector}
    \begin{barticle}
    \bauthor{\bsnm{Bongiorno}, \binits{C.}},
    \bauthor{\bsnm{Zhou}, \binits{Y.}},
    \bauthor{\bsnm{Kryven}, \binits{M.}},
    \bauthor{\bsnm{Theurel}, \binits{D.}},
    \bauthor{\bsnm{Rizzo}, \binits{A.}},
    \bauthor{\bsnm{Santi}, \binits{P.}},
    \bauthor{\bsnm{Tenenbaum}, \binits{J.}},
    \bauthor{\bsnm{Ratti}, \binits{C.}}:
    \batitle{Vector-based pedestrian navigation in cities}.
    \bjtitle{Nature Computational Science}
    \bvolume{1}(\bissue{10}),
    \bfpage{678}--\blpage{685}
    (\byear{2021})
    \end{barticle}
    \endbibitem
    
    \bibitem[\protect\citeauthoryear{Solera-Rico et~al.}{2024}]{solera2024beta}
    \begin{barticle}
    \bauthor{\bsnm{Solera-Rico}, \binits{A.}},
    \bauthor{\bsnm{Sanmiguel~Vila}, \binits{C.}},
    \bauthor{\bsnm{G{\'o}mez-L{\'o}pez}, \binits{M.}},
    \bauthor{\bsnm{Wang}, \binits{Y.}},
    \bauthor{\bsnm{Almashjary}, \binits{A.}},
    \bauthor{\bsnm{Dawson}, \binits{S.T.}},
    \bauthor{\bsnm{Vinuesa}, \binits{R.}}:
    \batitle{$\beta$-variational autoencoders and transformers for reduced-order modelling of fluid flows}.
    \bjtitle{Nature Communications}
    \bvolume{15}(\bissue{1}),
    \bfpage{1361}
    (\byear{2024})
    \end{barticle}
    \endbibitem
    
    \bibitem[\protect\citeauthoryear{Li et~al.}{2008}]{li2008public}
    \begin{botherref}
    \oauthor{\bsnm{Li}, \binits{Y.}},
    \oauthor{\bsnm{Perlman}, \binits{E.}},
    \oauthor{\bsnm{Wan}, \binits{M.}},
    \oauthor{\bsnm{Yang}, \binits{Y.}},
    \oauthor{\bsnm{Meneveau}, \binits{C.}},
    \oauthor{\bsnm{Burns}, \binits{R.}},
    \oauthor{\bsnm{Chen}, \binits{S.}},
    \oauthor{\bsnm{Szalay}, \binits{A.}},
    \oauthor{\bsnm{Eyink}, \binits{G.}}:
    A public turbulence database cluster and applications to study lagrangian evolution of velocity increments in turbulence.
    Journal of Turbulence
    (9),
    31
    (2008)
    \end{botherref}
    \endbibitem
    
    \bibitem[\protect\citeauthoryear{Sakurai and Napolitano}{2017}]{sakurai2017modern}
    \begin{bbook}
    \bauthor{\bsnm{Sakurai}, \binits{J.J.}},
    \bauthor{\bsnm{Napolitano}, \binits{J.}}:
    \bbtitle{Modern Quantum Mechanics},
    \bedition{2}nd edn.
    \bpublisher{Cambridge University Press},
    \blocation{Cambridge, UK}
    (\byear{2017})
    \end{bbook}
    \endbibitem
    
    \bibitem[\protect\citeauthoryear{Benzi et~al.}{1984}]{benzi1984multifractal}
    \begin{barticle}
    \bauthor{\bsnm{Benzi}, \binits{R.}},
    \bauthor{\bsnm{Paladin}, \binits{G.}},
    \bauthor{\bsnm{Parisi}, \binits{G.}},
    \bauthor{\bsnm{Vulpiani}, \binits{A.}}:
    \batitle{On the multifractal nature of fully developed turbulence and chaotic systems}.
    \bjtitle{Journal of Physics A: Mathematical and General}
    \bvolume{17}(\bissue{18}),
    \bfpage{3521}
    (\byear{1984})
    \end{barticle}
    \endbibitem
    
    \bibitem[\protect\citeauthoryear{Song and Kingma}{2021}]{song2021train}
    \begin{botherref}
    \oauthor{\bsnm{Song}, \binits{Y.}},
    \oauthor{\bsnm{Kingma}, \binits{D.P.}}:
    How to train your energy-based models.
    arXiv preprint arXiv:2101.03288
    (2021)
    \end{botherref}
    \endbibitem
    
    \bibitem[\protect\citeauthoryear{Wenliang et~al.}{2019}]{wenliang2019learning}
    \begin{bchapter}
    \bauthor{\bsnm{Wenliang}, \binits{L.}},
    \bauthor{\bsnm{Sutherland}, \binits{D.J.}},
    \bauthor{\bsnm{Strathmann}, \binits{H.}},
    \bauthor{\bsnm{Gretton}, \binits{A.}}:
    \bctitle{Learning deep kernels for exponential family densities}.
    In: \bbtitle{International Conference on Machine Learning},
    pp. \bfpage{6737}--\blpage{6746}
    (\byear{2019}).
    \bcomment{PMLR}
    \end{bchapter}
    \endbibitem
    
    \bibitem[\protect\citeauthoryear{Song and Ermon}{2019}]{song2019generative}
    \begin{botherref}
    \oauthor{\bsnm{Song}, \binits{Y.}},
    \oauthor{\bsnm{Ermon}, \binits{S.}}:
    Generative modeling by estimating gradients of the data distribution.
    Advances in neural information processing systems
    \textbf{32}
    (2019)
    \end{botherref}
    \endbibitem
    
    \bibitem[\protect\citeauthoryear{Roney and Ovchinnikov}{2022}]{roney2022state}
    \begin{barticle}
    \bauthor{\bsnm{Roney}, \binits{J.P.}},
    \bauthor{\bsnm{Ovchinnikov}, \binits{S.}}:
    \batitle{State-of-the-art estimation of protein model accuracy using alphafold}.
    \bjtitle{Physical Review Letters}
    \bvolume{129}(\bissue{23}),
    \bfpage{238101}
    (\byear{2022})
    \end{barticle}
    \endbibitem
    
    \bibitem[\protect\citeauthoryear{Barendregt}{1984}]{barendregt1984lambda}
    \begin{bbook}
    \bauthor{\bsnm{Barendregt}, \binits{H.P.}}:
    \bbtitle{The Lambda Calculus: Its Syntax and Semantics},
    \bedition{Revised 3rd} edn.,
    pp. \bfpage{23}--\blpage{315460}.
    \bpublisher{North-Holland},
    \blocation{Amsterdam}
    (\byear{1984})
    \end{bbook}
    \endbibitem
    
    \bibitem[\protect\citeauthoryear{Milner}{1978}]{MILNER1978348}
    \begin{barticle}
    \bauthor{\bsnm{Milner}, \binits{R.}}:
    \batitle{A theory of type polymorphism in programming}.
    \bjtitle{Journal of Computer and System Sciences}
    \bvolume{17}(\bissue{3}),
    \bfpage{348}--\blpage{375}
    (\byear{1978})
    \doiurl{10.1016/0022-0000(78)90014-4}
    \end{barticle}
    \endbibitem
    
    \bibitem[\protect\citeauthoryear{Parisi}{1981}]{parisi1981correlation}
    \begin{barticle}
    \bauthor{\bsnm{Parisi}, \binits{G.}}:
    \batitle{Correlation functions and computer simulations}.
    \bjtitle{Nuclear Physics B}
    \bvolume{180}(\bissue{3}),
    \bfpage{378}--\blpage{384}
    (\byear{1981})
    \end{barticle}
    \endbibitem
    
    \bibitem[\protect\citeauthoryear{Grenander and Miller}{1994}]{grenander1994representations}
    \begin{barticle}
    \bauthor{\bsnm{Grenander}, \binits{U.}},
    \bauthor{\bsnm{Miller}, \binits{M.I.}}:
    \batitle{Representations of knowledge in complex systems}.
    \bjtitle{Journal of the Royal Statistical Society: Series B (Methodological)}
    \bvolume{56}(\bissue{4}),
    \bfpage{549}--\blpage{581}
    (\byear{1994})
    \end{barticle}
    \endbibitem
    
    \end{thebibliography}
    

    \backmatter
    
    
    
    \section*{Funding}
    This work was supported in part by the National Key Research and Development Program of China (2022YFA1004100 [S.Y.]); in part by the National Natural Science Foundation of China (U21A6005 [S.Y.], 62172329 [X.R.]).
    
    \section*{Author contributions}
    
    S.Y. and X.R. conceived and supervised the project. H.Y., S.Y., and X.R. designed the study and algorithms. H.Y. implemented the code. H.Y., S.Y., X.R., and C.Z. performed the experiments. H.Y., S.Y., and X.R. wrote the manuscript. All authors discussed the results and reviewed the manuscript.
    
    \section*{Competing interests}
    
    The authors declare no competing interests.
    
    \bigskip
    
    \begin{table}[p!]
        \caption{Discovered expressions by Deflex in comparison with the ground truths}\label{tab1}%
    \begin{tabular}{@{}lll@{}}
    \toprule
        Rules &   & Ground truths and discovered by Deflex \\
        \midrule
    \multirow{2}{*}{Energy conservation}            & True       & $E = v^2 \cdot m / k_B T$                                                                                                                                      \\
                                                    & Discovered & $E = 0.992 v^2 \cdot m / k_B T $                                                                                                                                 \\ \hline
    \multirow{2}{*}{Momentum conservation}          & True       & $\sum{m_i \mathbf{v}_i} = C $                                                                                                                                 \\
                                                    & Discovered & $\sum{m_i \mathbf{v}_i} = C$                                                                                                                                  \\ \hline
    \multirow{2}{*}{\makecell{Maxwell-Boltzmann \\ distribution (MBD)}} & True       & $E = \frac{m v^2}{2 k_B T}$                                                                                                                                   \\
                                                    & Discovered & $E = \frac{0.511 m v^2}{k_B T}$                                                                                                                             \\ \hline
    \multirow{2}{*}{MBD with potential}             & True       & $E = \frac{m v^2 }{2k_B T} + \sum_{j \neq i} \left(\left\|\vec{x_j}-\vec{x_i}\right\|^{-12} - 2\left\|\vec{x_j}-\vec{x_i} \right\|^{-6}\right)$ \\
                                                    & Discovered & $E = \frac{0.506m v^2 }{k_B T} + \sum_{j \neq i} \left(0.97\left\|\vec{x_j}-\vec{x_i}\right\|^{-12} - 1.89\left\|\vec{x_j}-\vec{x_i} \right\|^{-6}\right)$ \\ \hline
    \multirow{2}{*}{Continuity of quantity}         & True       & $E = \lVert\nabla \cdot v\rVert \times \infty $                                                                                                               \\
                                                    & Discovered & $E = \lVert\nabla \cdot v\rVert \times 10^{26}$                                                                                                               \\ \hline
    \multirow{2}{*}{Newton viscosity law}           & True       & $\tau_{i j}=-\mu\left(\frac{\partial u_i}{\partial x_j}+\frac{\partial u_j}{\partial x_i}\right)$                                                             \\
                                                    & Discovered & $\tau_{i j}=-1.037\mu\left(\frac{\partial u_i}{\partial x_j}+\frac{\partial u_j}{\partial x_i}\right)$                                                        \\ \hline
    \multirow{2}{*}{Velocity distribution}          & True       & $E = \frac{1}{2} m v^2 + (\delta v)^P$                                                                                                                         \\
                                                    & Discovered & $E = 0.501 m v^2 + (\delta v)^{3.15}$                                                                                                                         \\ \hline
    \multirow{2}{*}{Navier-Stokes equation}         & True       & $E = \lVert \rho\left(\frac{\partial \mathbf{u}}{\partial t}+\mathbf{u} \cdot \nabla \mathbf{u}\right)+\nabla p-\mu \nabla^2 \mathbf{u}\rVert \times \infty $                                                                                 \\
                                                    & Discovered & $E = \lVert \rho\left(\frac{\partial \mathbf{u}}{\partial t}+\mathbf{u} \cdot \nabla \mathbf{u}\right)+\nabla p-\mu \nabla^2 \mathbf{u}\rVert \times 1.62\times 10^{21}$                                                                                       \\ \hline
    \multirow{2}{*}{Langevin}                       & True       & $E = \lVert4.468 \times 10^{16} \mathrm{~m} \dot{\boldsymbol{v}}+\boldsymbol{v}\rVert ^2$                                                                      \\
                                                    & Discovered & $E = \lVert4.438 \times 10^{16} m \dot{v}+1.003 v\rVert ^2$                                                                                                   \\ \hline
    \multirow{2}{*}{Lévy flight (Human)}            & True       & $E = D_1\log \Delta x$ \                                                                                                                                               \\
                                                    & Discovered & $E = 2.37\log \Delta x$                                                                                                                                              \\ \hline
    \multirow{2}{*}{Lévy flight (Avian)}            & True       & $E = D_2\log \Delta x$ \                                                                                                                                                   \\
                                                    & Discovered & $E = 1.627\log \Delta x$                                                                                                                                             \\ \hline
    \multirow{2}{*}{Power law}                      & True       & $E = \alpha \log l + \gamma l $                                                                                                                               \\
                                                    & Discovered & $E = 1.56 \log l + 0.00027 l$                                                                                                                                 \\ \hline
    \multirow{2}{*}{Crowd power law}                & True       & Undocumented                                                                                                                                                  \\
                                                    & Discovered & $E =1.63\log r_i-0.037 \log r_i\sum_{j \neq i} \mathbb{I}\left(\left\|x_t^{(j)}-x_t^{(i)}\right\|<2.46\right) $  \\                                              
                                        
    \botrule
    \end{tabular}
    \footnotesize $E$ denotes the energy value of the corresponding energy function. $k_B$ (Boltzmann constant), $\delta$, $\rho$, $P$, $D_1$, $D_2$, $\alpha$, and $\gamma$ are unprovided constants in the experiments.
    All other symbols (including temperature $T$ in the argon system) denote the provided features of observation data in the experiments.
    Detailed explanations of these features are described in SI.
    
    \end{table}
    
    \begin{table}[p!]
        \caption{Comparison of methods on discovery capability}\label{tab:forms}%
        \footnotesize
        \begin{tabular}{@{}lcccccccc@{}}
        \toprule
        Rules & Deflex & Operon & SciMED & SINDy & gplearn & AI Feynman & PySR & DEAP\\
        \midrule
        Energy conservation    & \checkmark& \checkmark & \checkmark & \checkmark & \checkmark & \checkmark & \checkmark & \checkmark \\    
        Momentum conservation    & \checkmark & \checkmark & \checkmark & \checkmark & \checkmark & \checkmark & \checkmark & \checkmark \\
        MBD  & \checkmark & \checkmark & \checkmark & N.A. & \checkmark & \checkmark & \checkmark & \checkmark \\
        MBD with potential & \checkmark & N.A. & N.A. & N.A. & \xmark & N.A. & N.A. & \xmark \\
        Continuity of quantity    & \checkmark & \checkmark & \checkmark & \checkmark & \checkmark & \checkmark & \checkmark & \checkmark \\
        Newton viscosity law    & \checkmark & \checkmark & \checkmark & \checkmark & \checkmark & \checkmark & \checkmark & \checkmark \\
        Velocity distribution    & \checkmark & \xmark & \xmark & N.A. & \xmark & \checkmark & \xmark & \xmark \\
        Navier-Stokes equation    & \checkmark & \checkmark\footnotemark & \xmark & \checkmark\footnotemark[\value{footnote}] & \xmark & \xmark & \xmark & \xmark \\
        Langevin    & \checkmark & \xmark & \xmark & N.A. & \xmark & \xmark & \xmark & \xmark \\
        Lévy flight (Human)    & \checkmark & \xmark & \xmark & N.A. & \xmark & \checkmark & \checkmark & \checkmark \\
        Lévy flight (Avian)    & \checkmark & \xmark & \xmark & N.A. & \xmark & \checkmark & \checkmark & \checkmark \\
        Power law   & \checkmark & \xmark & \xmark & N.A. & \xmark & \checkmark & \checkmark & \checkmark \\
        \bottomrule
        \end{tabular}
        \footnotetext{Operon and SINDy can recover a 2D Navier--Stokes form (cylinder wake) but fail in the 3D JHTDB setting; moreover, their recovered 2D forms do not include pressure-field terms (i.e., $\nabla p$).}
        \footnotesize \checkmark ~denotes a successful discovery, \xmark ~denotes a failure, and N.A. denotes an unsupported task (i.e., not applicable).
    \end{table}
    
    \end{document}